\documentclass{article}
\usepackage[accepted]{icml2020}

\usepackage[utf8]{inputenc} 
\usepackage[T1]{fontenc}    
\usepackage[colorlinks = true, citecolor=black]{hyperref}       
\usepackage{url}            
\usepackage{booktabs}       
\usepackage{amsfonts}       
\usepackage{nicefrac}       
\usepackage{microtype}      
\usepackage{subfigure}
\usepackage{graphicx}
\usepackage{enumitem}
\usepackage{titlesec}
\usepackage{wrapfig}
\usepackage[small]{caption}
\usepackage{chngcntr}

\usepackage{dsfont}
\usepackage{amsthm}
\usepackage{amsmath}
\usepackage{amssymb}
\usepackage{color}
\usepackage{xspace}
\usepackage{graphicx}
\usepackage{pgfplots, pgfplotstable}
\usepackage{xcolor,colortbl}

\def\smath#1{\text{\scalebox{.9}{$#1$}}}

\def\sfrac#1#2{\smath{\frac{#1}{#2}}}

\renewcommand{\mid}[0]{\:\vert\:}

\def\Qt{Q_{\theta}}
\def\Qtp{Q_{\thetaP}}
\def\ld{\ell_{\lambda}}

\def\expected{\mathbb{E}}

\newcommand{\R}[1]{R}
\newcommand{\G}[1]{G}

\newcommand{\Ad}[1]{A}

\newcommand{\CC}[1]{C}

\def\setA{\mathcal{A}}
\def\setD{\mathcal{D}}

\def\setS{\mathcal{S}}

\def\setB{\mathcal{B}}
\def\thetaP{\theta^{\prime}}

\def\calL{\mathcal{L}}

\def\Real{\mathbb{R}}

\def\@onedot{\ifx\@let@token.\else.\null\fi\xspace}
\def\etc{{\em etc}}
\def\eg{{\em e.g.,}\xspace}
\def\ie{{\em i.e.,}\xspace}
\def\wrt{w.r.t.\xspace}  

\def\versus{{\em vs.}\xspace}

\newcommand{\figref}[1]{Figure~\ref{#1}}

\newcommand{\tabref}[1]{Table~\ref{#1}}

\usepackage{amsmath,amsfonts,bm}

\def\1{\bm{1}}

\def\ermK{{\textnormal{K}}}

\def\ermP{{\textnormal{P}}}

\DeclareMathAlphabet{\mathsfit}{\encodingdefault}{\sfdefault}{m}{sl}
\SetMathAlphabet{\mathsfit}{bold}{\encodingdefault}{\sfdefault}{bx}{n}

\def\title{An Optimistic Perspective on Offline Reinforcement Learning}

\icmltitlerunning{\title}

\begin{document}

\twocolumn[
\icmltitle{\title}



\icmlsetsymbol{aires}{*}

\begin{icmlauthorlist}
\icmlauthor{Rishabh Agarwal}{goo}
\icmlauthor{Dale Schuurmans}{goo,ab}
\icmlauthor{Mohammad Norouzi}{goo}
\end{icmlauthorlist}

\icmlaffiliation{goo}{Google Research, Brain Team}
\icmlaffiliation{ab}{University of Alberta}

\icmlcorrespondingauthor{Rishabh Agarwal}{rishabhagarwal@google.com}
\icmlcorrespondingauthor{Mohammad Norouzi}{mnorouzi@google.com}

\icmlkeywords{Offline RL, Reinforcement Learning, Atari games, Batch RL, Deep RL}

\vskip 0.3in
]



\printAffiliationsAndNotice{} 

\vspace{-0.4cm}
\begin{abstract}
Off-policy reinforcement learning~(RL) using a fixed offline dataset
of logged interactions is an important consideration in real world
applications.
This paper studies offline RL using the DQN Replay Dataset
comprising the entire replay experience of a DQN agent on 60 Atari
2600 games.
We demonstrate that recent off-policy deep RL algorithms,
even when trained solely on this fixed dataset, outperform the fully-trained DQN agent.
To enhance generalization in the offline setting,
we present Random Ensemble Mixture (REM),
a robust $Q$-learning algorithm that enforces optimal Bellman consistency on
random convex combinations of multiple $Q$-value estimates.
Offline REM trained on the DQN Replay Dataset surpasses
strong RL baselines.  Ablation studies highlight the role of offline dataset size and diversity as well as the algorithm choice in  our positive results. Overall, the results here present an
optimistic view that robust RL algorithms used on sufficiently
large and diverse offline datasets can lead to high quality policies.
To provide a testbed for offline RL and reproduce our results, the DQN Replay Dataset is released at
\href{https://offline-rl.github.io}{offline-rl.github.io}.
\end{abstract}

\vspace{-0.4cm}
\section{Introduction}


One of the main reasons behind the success of deep learning
is the availability of large and diverse datasets such as
ImageNet~\citep{imagenet_cvpr09} to train expressive deep neural
networks. By contrast, most reinforcement learning~(RL)
algorithms~\citep{sutton2018reinforcement} assume that an agent
interacts with an online environment or simulator and learns from its
own collected experience. This limits online RL's applicability to complex real world problems, where active data
collection means gathering large amounts of diverse
data from scratch
per experiment, which can be expensive,
unsafe, or require a
high-fidelity simulator that is often difficult to build~\citep{dulac2019challenges}.

Offline RL concerns the problem of learning a policy from a fixed dataset of trajectories, without any further interactions with the environment. This setting
can leverage the vast amount of existing logged interactions for
real world decision-making problems such as
robotics~\citep{cabi2019framework, dasari2019robonet}, autonomous
driving~\citep{yu2018bdd100k}, recommendation
systems~\citep{strehletal10, bottou2013counterfactual}, and
healthcare~\citep{shortreed2011informing}. The effective use of
such datasets would not only make real-world RL
more practical, but would also enable better generalization
by incorporating diverse prior experiences.

In offline RL, an agent does not receive any new corrective feedback
from the online environment and needs to generalize from a fixed dataset of
interactions to new online interactions during evaluation. In
principle, off-policy algorithms can learn from data collected by any
policy, however, recent work~\citep{fujimoto2018off,
kumar2019stabilizing, wu2019behavior, siegel2020keep} presents a
discouraging view that standard off-policy deep RL algorithms diverge
or otherwise yield poor performance in the offline setting. Such
papers propose remedies by regularizing the learned policy to stay
close to the training dataset of offline trajectories.
Furthermore, \citet{Zhang2017ADL} assert that a large replay buffer can
even hurt the performance of off-policy algorithms due to its
``off-policyness''.

\begin{figure*}[t]
  \begin{center}
    \footnotesize
    \begin{tabular}{@{}c@{\hspace*{.3cm}}c@{}}
      \includegraphics[width=0.49\linewidth]{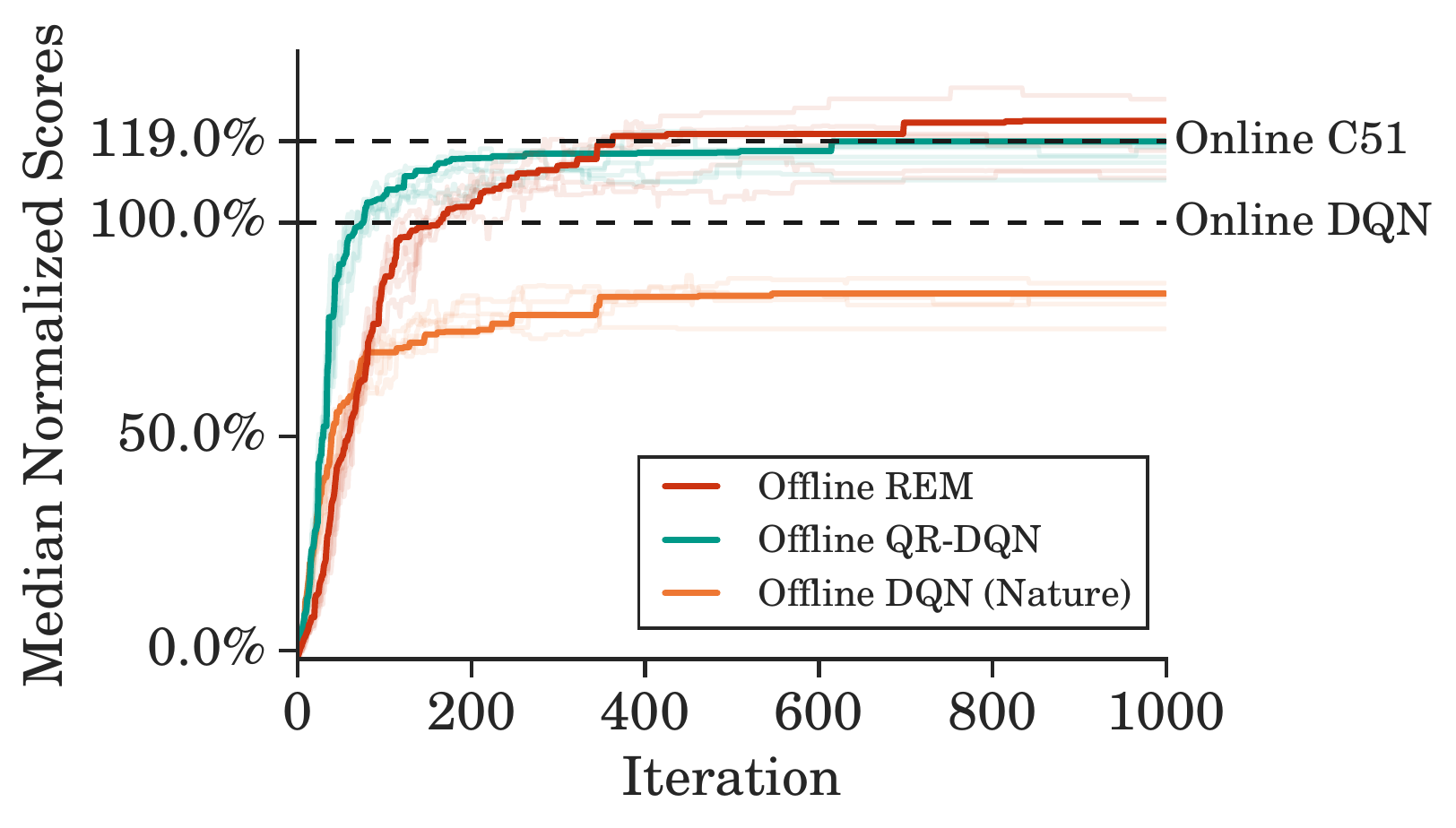} & \includegraphics[width=0.45\linewidth]{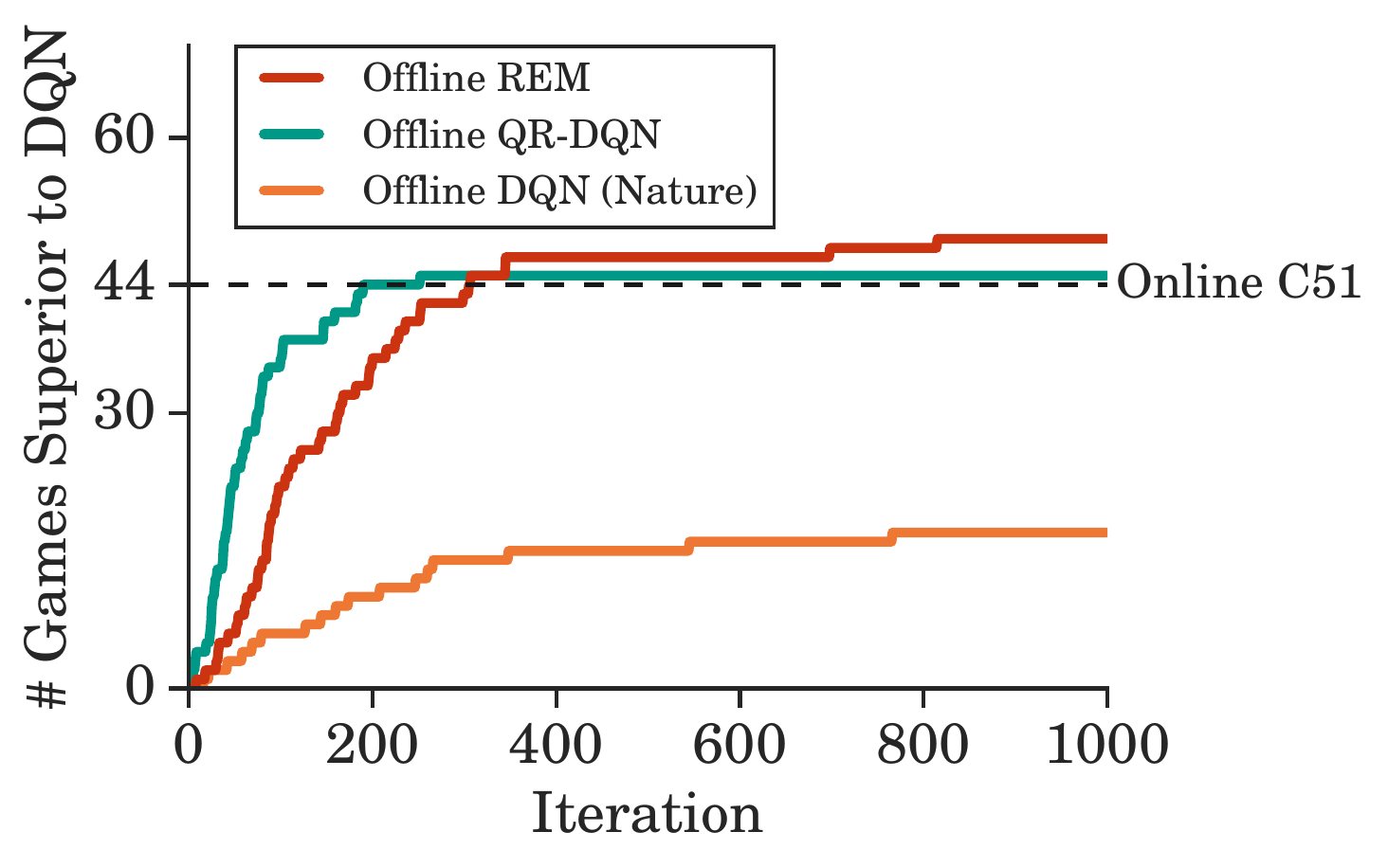} \\
      (a) & (b)
    \end{tabular}
    \vspace{-0.15cm}
    \caption{{\bf Offline RL on Atari 2600.} (a) Median normalized evaluation scores averaged over 5 runs~(shown as traces) across stochastic version of 60 Atari 2600 games of offline agents trained using the DQN replay dataset.
    (b) Number of games where an offline agent achieves a higher score than fully-trained DQN~(Nature) as a function of training iterations. Each iteration corresponds to 1 million training frames. Offline REM outperforms offline QR-DQN and
    DQN~(Nature). The comparison with online C51 gives a sense to the reader about the magnitude of the improvement from offline agents over the best policy in the entire DQN replay dataset.}\label{fig:batch_median}
  \end{center}
\vspace{-0.45cm}
\end{figure*}

By contrast, this paper presents an optimistic perspective on offline
RL that with sufficiently large and diverse datasets, robust RL
algorithms, without an explicit correction for distribution mismatch,
can result in high quality policies.
The contributions of this paper can be summarized as:
\begin{itemize}[topsep=0pt, partopsep=0pt, leftmargin=13pt, parsep=0pt, itemsep=4pt]
\item An offline RL setup is proposed for evaluating algorithms on Atari 2600 games~\citep{bellemare2013arcade}, based
on the logged replay data of a DQN agent~\citep{mnih2015human} comprising $50$ million 
(observation, action, reward, next observation) tuples per game.
This setup reduces the computation cost of the experiments considerably and helps improve reproducibility 
by standardizing training using a fixed dataset. The DQN Replay Dataset and our code\footnote{Open-source code at \href{https://github.com/google-research/batch_rl}{ github.com/google-research/batch\_rl}.} is released to enable
offline optimization of RL algorithms on a common ground.
\item Contrary to recent work, we show that recent off-policy RL algorithms trained solely on offline data can be successful.
For instance, offline QR-DQN~\citep{dabney2018distributional} trained on the DQN replay dataset outperforms the \textit{best} policy in the DQN replay dataset. This discrepancy is attributed to the differences in offline dataset size and diversity as well as the choice of RL algorithm.
\item A robust $Q$-learning algorithm called {\em Random Ensemble
Mixture~(REM)} is presented, which enforces optimal Bellman
consistency on random convex combinations of multiple $Q$-value
estimates. Offline REM shows strong generalization performance in the offline setting,
and outperforms offline QR-DQN. The comparison with
online C51~\citep{bellemare2017distributional}, a strong RL baseline illustrates the
relative size of the gains from exploitation of the logged DQN data with REM.

\end{itemize}

\vspace{-0.1cm}
\section{Off-policy Reinforcement Learning}

An interactive environment in reinforcement learning~(RL) is typically described as a Markov
decision process (MDP) $(\setS, \setA, R,
P, \gamma)$ \citep{puterman1994markov}, with a state space $\setS$, an
action space $\setA$, a stochastic reward function $R(s, a)$,
transition dynamics $P(s' | s, a)$ and a discount factor $\gamma \in
[0, 1)$. A stochastic policy $\pi(\cdot\mid s)$ maps each state
$s \in \setS$ to a distribution (density) over actions.

For an agent following the policy $\pi$, the action-value
function, 
denoted $Q^\pi(s, a)$, is defined as the expectation of cumulative discounted
future rewards, \ie
\begin{align}
Q^\pi(s, a) :={}& \expected\left[ \sum\nolimits_{t=0}^\infty \gamma^t R(s_t, a_t) \right], \\
s_0 = s, a_0 =&~a, s_t \sim P(\cdot \mid s_{t-1}, a_{t-1}), a_t \sim \pi(\cdot \mid s_{t}). \nonumber 
\end{align}
The goal of RL is to find an optimal policy $\pi^*$
that attains maximum expected return,
for which $Q^{\pi^*}(s, a) \geq Q^{\pi}(s, a)$ for all $\pi, s, a$.
The Bellman optimality equations~\citep{bellman1957dynamic} characterize
the optimal policy in terms of the optimal $Q$-values, denoted
$Q^* = Q^{\pi^*}$, via:
\begin{equation}
Q^*(s, a) = \expected\ R(s, a) + \gamma \expected_{s'\sim P} \max_{a' \in \setA} Q^*(s', a')~.
\label{eq:q*}
\end{equation}
To 
learn a policy from interaction with the environment,
$Q$-learning~\citep{watkins1992q} iteratively improves an
approximate estimate of $Q^*$, denoted $Q_{\theta}$, by repeatedly
regressing the LHS of \eqref{eq:q*} to target values defined by samples from the RHS of \eqref{eq:q*}.
For large and complex state spaces, approximate $Q$-values are obtained
using a neural network as the function approximator.
To further stabilize optimization, a target network $Q_{\thetaP}$ with frozen
parameters may be used for computing the learning target~\citep{mnih2013playing}. The target
network parameters $\thetaP$ are updated to the current $Q$-network
parameters $\theta$ after a fixed number of time
steps.

DQN~\citep{mnih2013playing, mnih2015human} parameterizes $Q_\theta$ with a
convolutional neural network~\citep{lenet} and uses $Q$-learning with a target network
while following an $\epsilon$-greedy policy with respect to $Q_\theta$ for data
collection. DQN minimizes the temporal difference~(TD) error $\Delta_\theta$ using the loss $\calL(\theta)$ on
mini-batches of agent's past experience tuples, $(s, a, r, s')$,
sampled from an experience replay buffer $\setD$~\citep{lin1992self}
collected during training:
\begin{align}\label{eq:dqn_loss}
\calL(\theta) ={}& \expected_{s, a, r, s' \sim \setD} \left[ \ld \left( \Delta_\theta(s, a, r, s') \right)\right],\\
\Delta_\theta(s, a, r, s') &=~\Qt(s, a) - r - \gamma\max_{a'} \Qtp(s', a') \nonumber
\end{align}

where $l_{\lambda }$ is the Huber loss~\citep{huber1964robust} given by
\begin{align}
    \ld(u) = \begin{cases}
        \frac{1}{2} u^2,\quad \ &\text{if } |u| \le \lambda\\
        \lambda(|u| - \frac{1}{2}\lambda),\quad \ &\text{otherwise.}
    \end{cases}
\end{align}
$Q$-learning is an \textit{off-policy}
algorithm~\citep{sutton2018reinforcement} since the learning target
can be computed without any consideration of how the experience was
generated.

A family of recent off-policy deep RL algorithms, which serve as a strong
baseline in this paper, include Distributional
RL~\citep{bellemare2017distributional, jaquette1973markov} methods. Such
algorithms estimate a density over returns for each state-action
pair, denoted $Z^{\pi}(s, a)$, instead of directly estimating the mean
$Q^{\pi}(s, a)$.
Accordingly, one can express a form of distributional Bellman optimality 
as
\begin{align}
Z^{*}(s, a) \stackrel{D}{=} r +
\gamma Z^{*}(s', \mathrm{argmax}_{a' \in \setA}~Q^*(s',a')), \label{eq:distributional_bellman}\\
\mathrm{where}~r \sim R(s, a),\ s' \sim P(\cdot \mid s, a). \nonumber
\end{align}
and
$\stackrel{D}{=}$ denotes distributional 
equivalence
and
$Q^*(s',a')$ is estimated by taking an expectation with respect to
$Z^{*}(s', a')$. C51~\citep{bellemare2017distributional} approximates
$Z^{*}(s, a)$ by using a categorical distribution over a set of pre-specified anchor points,
and distributional QR-DQN~\citep{dabney2018distributional}
approximates the return density by using a uniform mixture of $K$ Dirac delta functions, \ie

\begin{equation*}
Z_\theta(s, a) := \frac{1}{K} \sum_{i=1}^K \delta_{\theta_i(s,
a)},\ Q_{\theta}(s, a)
= \frac{1}{K} \sum_{i=1}^K \theta_i(s, a).
\end{equation*}

QR-DQN 
outperforms C51 and DQN and obtains state-of-the-art results on
Atari 2600 games, among agents that do not exploit $n$-step
updates~\citep{sutton1988learning} and prioritized replay~\citep{schaul2015prioritized}. This paper avoids using
$n$-step updates and prioritized replay to keep the empirical study
simple and focused on deep $Q$-learning algorithms.

\vspace{-0.1cm}
\section{Offline Reinforcement Learning}

\begin{figure*}[t]
\begin{center}
\includegraphics[width=0.85\linewidth]{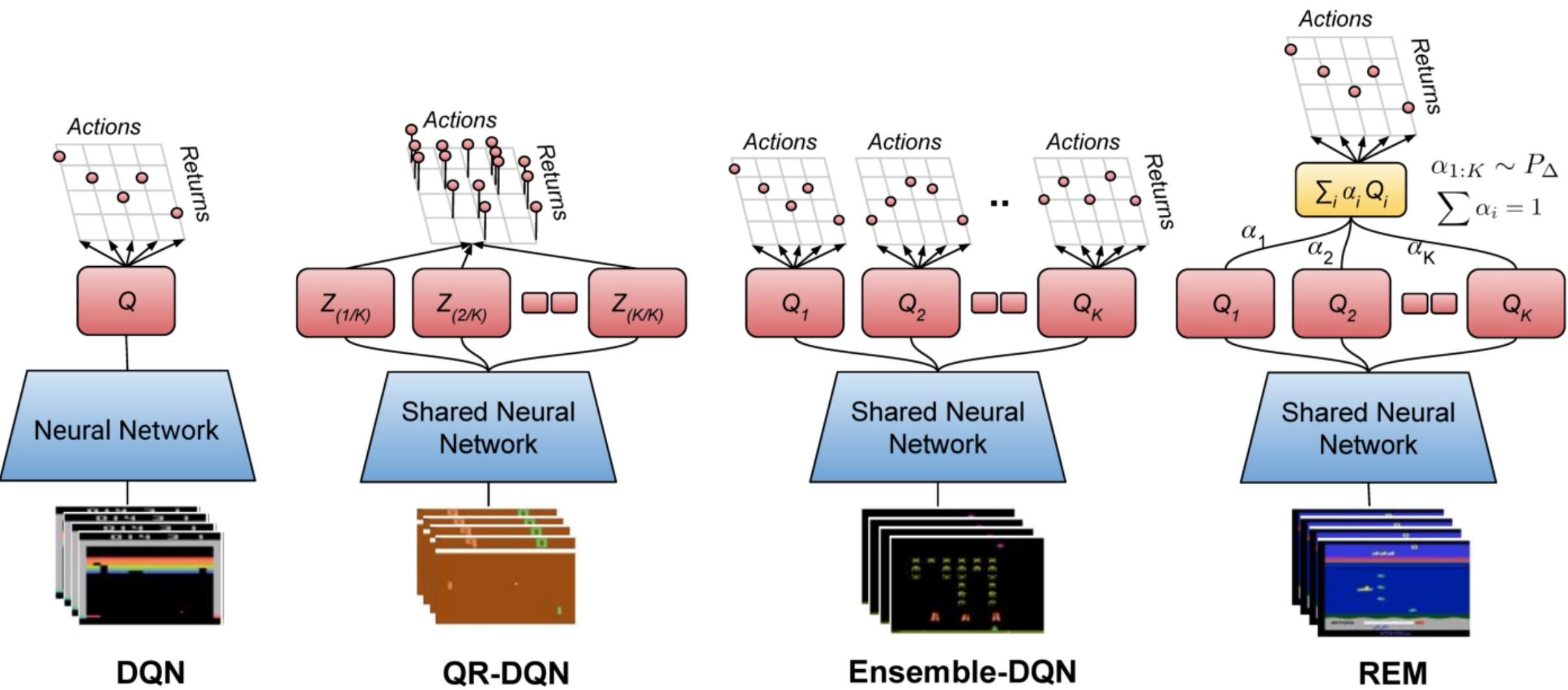}
\end{center}
\vspace{-0.14cm}
\caption{Neural network architectures for DQN, distributional QR-DQN and the proposed expected RL variants, \ie\ Ensemble-DQN and REM,
with the same multi-head architecture as QR-DQN. The individual $Q$-heads share all of the neural network layers except the
final fully connected layer.
In QR-DQN, each head~(red rectangles) corresponds to a specific quantile of the return distribution, while in the proposed variants, each head approximates the optimal $Q$-function.}\label{fig:architechture}
\vspace{-0.3cm}
\end{figure*}

Modern off-policy deep RL algorithms~(as discussed above) perform remarkably well on common
benchmarks such as the Atari 2600 games~\citep{bellemare2013arcade}
and continuous control MuJoCo tasks~\citep{todorov2012mujoco}.  Such
off-policy algorithms
are considered ``online'' because they alternate between optimizing a
policy and using that policy to collect more data.  Typically, these
algorithms keep a sliding window of most recent experiences in a
finite replay buffer~\citep{lin1992self}, throwing away stale data to
incorporate most fresh~(on-policy) experiences.

Offline RL, in contrast to online RL, describes the fully off-policy setting of
learning using a fixed dataset of experiences, without any further interactions with the environment.
We advocate the use of offline RL to help isolate an RL algorithm's ability to {\em exploit} experience and generalize \versus
its ability to {\em explore} effectively.
The offline RL setting removes design choices related to the
replay buffer and exploration; therefore, it is simpler to
experiment with and reproduce than the online setting.

Offline RL is considered challenging due to the \textit{distribution mismatch} between the current policy
and the offline data collection policy, \ie when the policy being learned takes a different action than the data collection policy, we don't know the reward that should be provided. This paper revisits offline RL and investigates whether
off-policy deep RL agents trained solely on offline data can be
successful without correcting for distribution mismatch.

\vspace{-0.1cm}
\section{Developing Robust Offline RL Algorithms}

In an online RL setting, an agent can acquire on-policy data from the environment, which ensures a virtuous cycle where the agent chooses actions that it thinks will lead to
high rewards and then receives feedback to correct its errors. Since it is not possible to collecting additional data in the offline RL setting, it is necessary to reason
about generalization using the fixed dataset.
We investigate whether one can design robust RL algorithms with an emphasis on improving generalization in the offline setting.
Ensembling is commonly used in supervised learning to improve generalization.
In this paper, we study two deep $Q$-learning algorithms, Ensemble DQN and REM, which adopt ensembling, to improve stability.

\subsection{Ensemble-DQN}
Ensemble-DQN is a simple extension of DQN that approximates the
$Q$-values via an ensemble of parameterized $Q$-functions~\citep{fausser2015neural,
osband2016deep, anschel2017averaged}.
Each $Q$-value estimate,
denoted $Q_{\theta}^{k}(s, a)$, is trained against its own target
$Q_{\thetaP}^{k}(s, a)$, similar to Bootstrapped-DQN~\citep{osband2016deep}.
The $Q$-functions
are optimized using identical mini-batches in the same order, starting from
different parameter initializations. The loss $\calL(\theta)$ takes the form,
\begin{align}\label{ensemble_eq}
\calL(\mathrm{\theta}) ={}& \sfrac{1}{K} \sum_{k=1}^{K} \expected_{s, a, r, s' \sim \setD} \left[ \ld\left(\Delta^{k}_\theta(s, a, r, s') \right)\right],\\
\Delta^{k}_\theta(s, a, r, s') &= \Qt^{k}(s, a) - r - \gamma\max_{a'} \Qtp^{k}(s', a') \nonumber
\end{align}

where $l_{\lambda }$ is the Huber loss. While Bootstrapped-DQN uses one
of the $Q$-value estimates in each episode to improve exploration, in
the offline setting, we are only concerned with the ability of
Ensemble-DQN to exploit better and use the mean of the $Q$-value estimates for evaluation.

\subsection{Random Ensemble Mixture~(REM)}\label{sec:rem}

Increasing the number of models used for ensembling typically improves
the performance of supervised learning
models~\citep{shazeer2017outrageously}.  This raises the question
whether one can use an ensemble over an exponential number of
$Q$-estimates in a computationally efficient manner. Inspired by
dropout~\citep{srivastava2014dropout}, we propose Random Ensemble
Mixture for off-policy RL.

Random Ensemble Mixture (REM) uses multiple parameterized
$Q$-functions to estimate the $Q$-values, similar to Ensemble-DQN.
The key insight behind REM is that one can think of a convex
combination of multiple $Q$-value estimates as a $Q$-value
estimate itself. This is especially true at the fixed point, where all
of the $Q$-value estimates have converged to an identical
$Q$-function. Using this insight, we train a family of $Q$-function
approximators defined by mixing probabilities on a $(K-1)$-simplex.

Specifically, for each mini-batch, we randomly draw a categorical
distribution $\alpha$, which defines a convex combination of the $K$
estimates to approximate the optimal $Q$-function. This approximator
is trained against its corresponding target to minimize the TD
error. The loss $\calL(\theta)$ takes the form,
\begin{align}\label{eq:sqn}
\calL(\theta) ={}& \expected_{s, a, r, s' \sim \setD} \left[\expected_{\alpha \sim \ermP_{\Delta}} \left[\ld \big( \Delta^{\alpha}_\theta(s, a, r, s') \big)\right]\right], \\
\Delta^{\alpha}_\theta =& \sum_{k} \alpha_{k} \Qt^{k}(s, a) - r - \gamma\max_{a'} \sum_{k} \alpha_{k}\Qtp^{k}(s', a') \nonumber
\end{align}
where $\ermP_{\Delta}$ represents a probability distribution over the
standard $(K-1)$-simplex $\Delta^{\ermK - 1}
= \{\alpha \in \Real^{\ermK}: \alpha_{1}+ \alpha_{2} + \dots
+ \alpha_{K}=1, \alpha_{k}\geq 0, k=1,\dots,K\}$.

REM considers $Q$-learning as a constraint satisfaction problem based
on Bellman optimality constraints~\eqref{eq:q*} and $\calL(\theta)$ can be viewed as an infinite set of constraints
corresponding to different mixture probability distributions.
For action selection, 
we use the average of the $K$ value estimates as the $Q$-function, \ie~$Q(s,a)
= \sum_{k} \Qt^{k}(s, a) / K$. REM is easy to implement and analyze~(see Proposition 1),
and can be viewed as a simple regularization technique for value-based RL.
In our experiments, we use a very
simple distribution $\ermP_{\Delta}$: we first
draw a set of $K$ values {\em i. \!i.\! d.} from Uniform~(0, 1) and
normalize them to get a valid categorical
distribution, \ie~$\alpha'_k \sim U(0, 1)$ followed by $\alpha_k
= \alpha'_k / \sum \alpha'_i$.

\textbf{Proposition 1}.
{\it
Consider the assumptions:
(a) The distribution $\ermP_{\Delta}$ has full support over the entire $(K-1)$-simplex.
(b) Only a finite number of distinct $Q$-functions globally minimize the loss in \eqref{eq:dqn_loss}.
(c) $Q^*$ is defined in terms of the MDP induced by the data distribution $\mathcal{D}$.
(d) $Q^*$ lies in the family of our function approximation.
Then, at the global minimum of $\calL({\theta})$~\eqref{eq:sqn} for a multi-head $Q$-network:
\setlist{nolistsep}
\begin{enumerate}[label=(\roman*)]
\itemsep0.14em
\item Under assumptions (a) and (b), all the $Q$-heads represent identical $Q$-functions.
\item Under assumptions (a)--(d), the common global solution is $Q^*$.
\end{enumerate}
}

The proof of \textit{(ii)} follows from \textit{(i)} and the fact that \eqref{eq:sqn} is lower bounded
by the TD error attained by $Q^*$. The proof of part \textit{(i)} can be found
in the supplementary material.

\vspace{-0.1cm}
\section{Offline RL on Atari 2600 Games}
\label{sec:offline_results}

\begin{figure*}[t]
  \begin{center}
        \includegraphics[width=\linewidth]{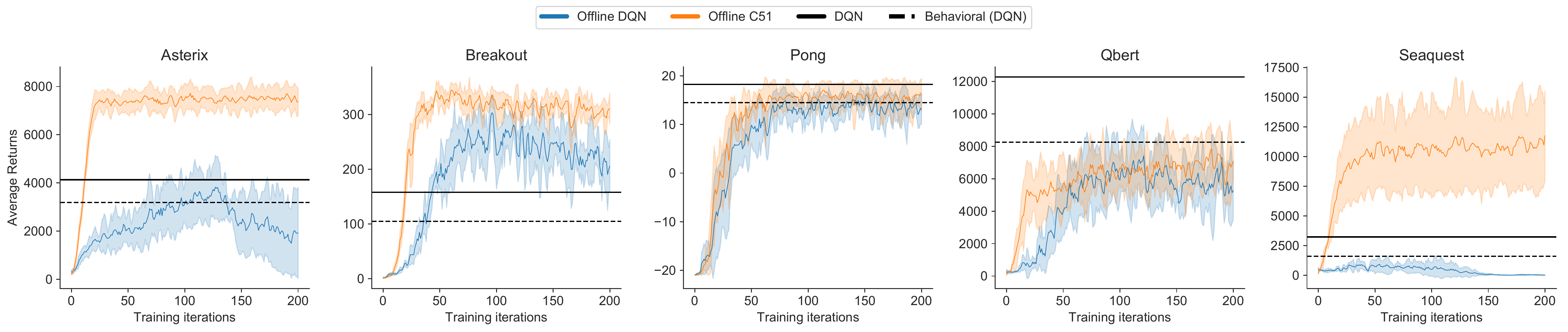}
        \vspace{-0.4cm}
        \caption{{\bf Offline Agents on DQN Replay Dataset.}
        Average online scores of C51 and DQN~(Nature) agents trained offline on 5 games with sticky actions for the same number of gradient steps as the online DQN agent.
        The scores are averaged over 5 runs (shown as traces) and smoothed over a sliding window of 5 iterations and error bands show standard deviation. The solid and dashed horizontal line show the performance of best policy~(\ie fully-trained DQN) and the average behavior policy in the DQN Replay Dataset respectively.}\label{fig:c51_vs_dqn_batch}
  \end{center}
\vspace{-0.05cm}
\end{figure*}

\begin{figure*}[t]
  \begin{center}
    \footnotesize
    \begin{tabular}{@{}c@{}c@{}}
      \includegraphics[width=0.5\linewidth]{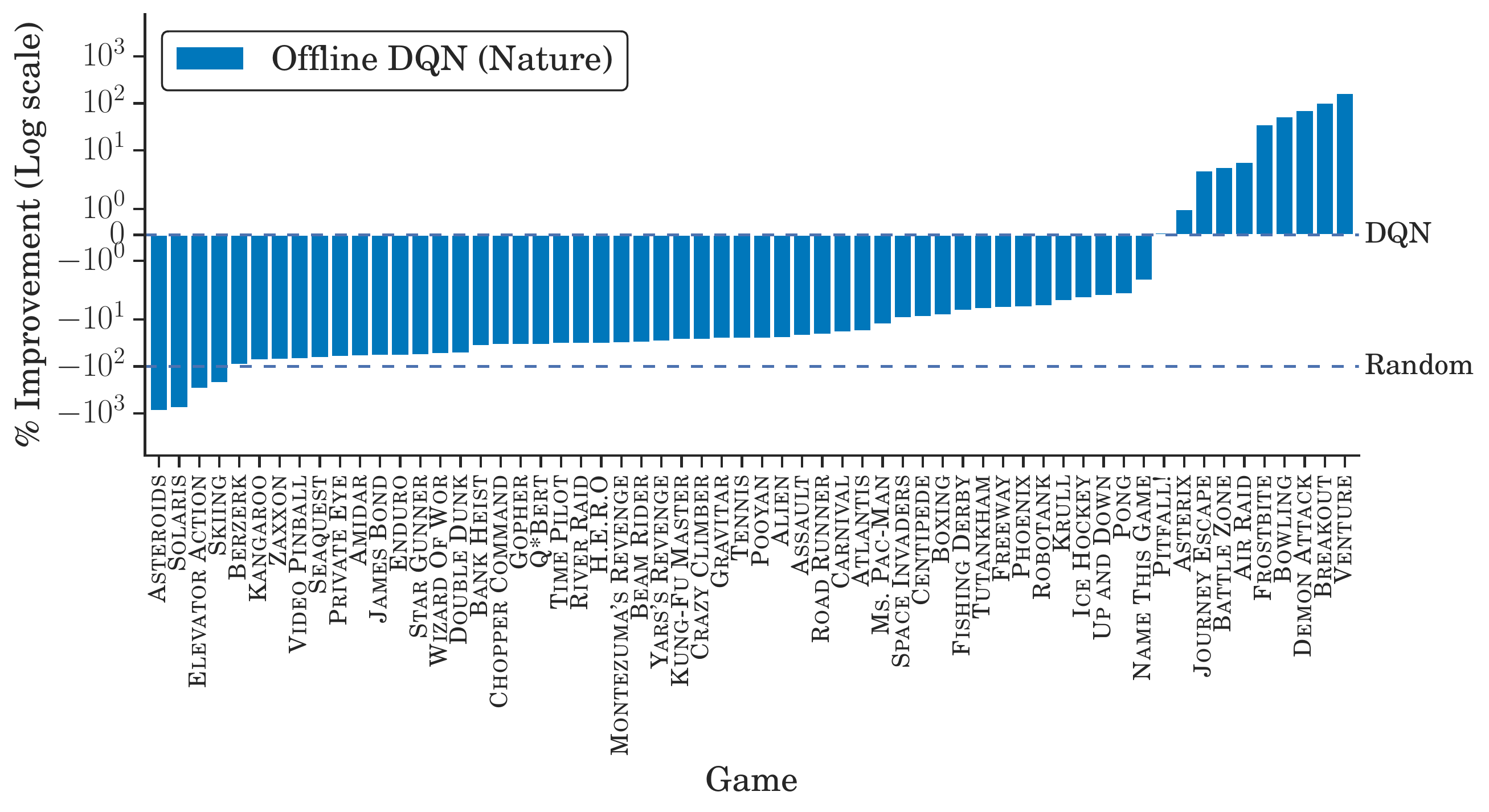} & \includegraphics[width=0.5\linewidth]{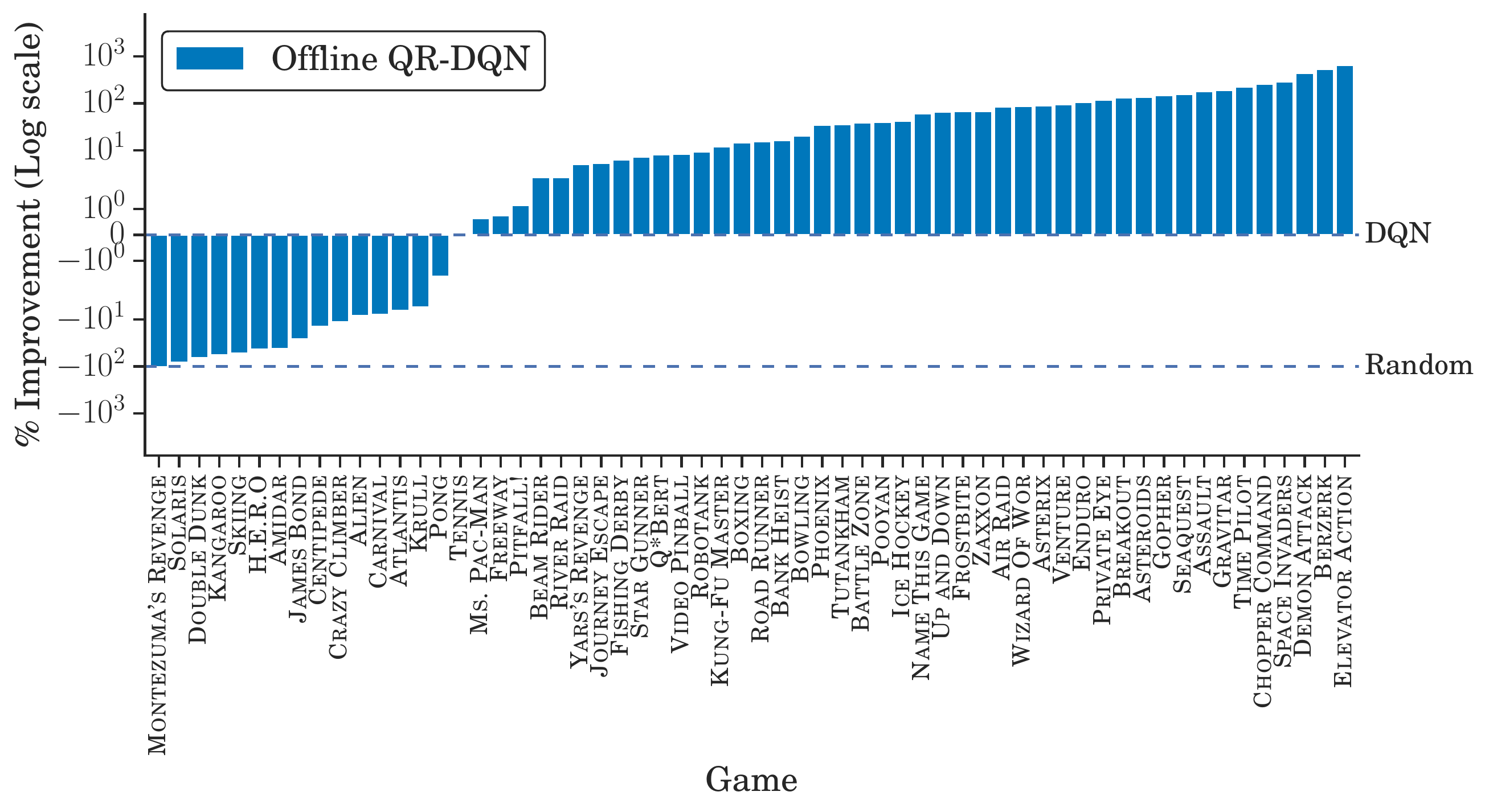} \\
      (a) & (b)
    \end{tabular}
    \vspace{-0.1cm}
    \caption{{\bf Offline QR-DQN \versus DQN~(Nature).} Normalized performance improvement (in \%) over fully-trained online DQN~(Nature), per game, of (a) offline DQN~(Nature) and (b) offline QR-DQN
    trained using the DQN replay dataset for same number of gradient updates as online DQN. The normalized online score for each game is 100\% and 0\% for online DQN and random agents respectively.}\label{fig:batch_dqn}
  \end{center}
\vspace{-0.4cm}
\end{figure*}

To facilitate a study of offline RL, we create the DQN Replay Dataset by training
several instances of DQN~(Nature) agents~\citep{mnih2015human} on 60 Atari
2600 games for 200 million frames each, with a frame skip of
4~(standard protocol) and sticky actions~\citep{machado2018revisiting}~(with 25\% probability, the agent's previous action is executed instead of the current action). On each game, we train $5$
different agents with random initialization, and store all of the
tuples of {(observation, action, reward, next observation)}
encountered during training into $5$ replay datasets per game resulting in a total of 300 datasets. 

Each game replay dataset is approximately 3.5 times larger than ImageNet~\citep{imagenet_cvpr09} and include samples from all of the intermediate (diverse) policies seen during the optimization of the online DQN agent. \figref{fig:dqn_vs_batch} shows the learning curves of the individual agents used for data collection as well as the performance of best policy found during training~(which we use for comparison).

{\bf Experiment Setup}. The DQN Replay Dataset is used for training
RL agents, offline, without any interaction with the
environment during training. We use the hyperparameters provided in Dopamine
baselines~\citep{castro2018dopamine} for a standardized
comparison~(Appendix~\ref{sec:hyperparameters}) and report game scores using
a normalized scale~(Appendix~\ref{sec:score_normalization}). Although the game replay datasets contain data collected by a DQN agent improving over time as training progresses, we compare the performance of offline agents against the best performing agent obtained after training~(\ie fully-trained DQN). 
The evaluation of the offline agents is done online for a limited number of times in the intervals of 1 million training frames. For each game, we evaluate the 5 offline agents trained~(one per dataset), using online returns, reporting the best performance averaged across the 5 agents. 

\subsection{Can standard off-policy RL algorithms with no environment interactions succeed?}\label{sec:part1_experiments}

Given the logged replay data of the DQN agent, it is natural to ask
how well an offline variant of DQN solely trained using this dataset
would perform?
Furthermore, whether more recent
off-policy algorithms are able to exploit the DQN Replay Dataset more
effectively than offline DQN. To investigate these questions, we train
DQN~(Nature) and QR-DQN agents, offline, on the DQN
replay dataset for the same number of gradient updates as online DQN.

Figure~\ref{fig:batch_dqn} shows that offline DQN underperforms fully-trained online DQN on
all except a few games where it achieves much higher scores than
online DQN with the same amount of data and gradient updates.
Offline QR-DQN, on the other hand, outperforms offline DQN and online DQN on most of the
games~(refer to Figure~\ref{fig:dqn_vs_batch} for learning curves).
Offline C51 trained using the DQN Replay Dataset also considerably improves upon offline
DQN~(Figure~\ref{fig:c51_vs_dqn_batch}). Offline QR-DQN outperforms offline C51. 

These results demonstrate that it is possible to optimize strong Atari agents offline using standard deep RL algorithms on DQN Replay Dataset
without constraining the learned policy to stay close to the training dataset of offline trajectories.
Furthermore, the disparity between the performance of offline QR-DQN/C51 and DQN~(Nature)
indicates the difference in their ability to exploit offline data.

\subsection{Asymptotic performance of offline RL agents}

In supervised learning, asymptotic performance matters more than performance within a fixed budget of gradient updates.
Similarly, for a given sample complexity, we prefer RL algorithms that perform the best as long as the number of gradient updates
is feasible. Since the sample efficiency for an offline dataset is fixed, we train offline agents for 5 times as many gradient updates as DQN.

\begin{table}[t]
    \caption{\textbf{Asymptotic performance of offline agents}. Median normalized scores~(averaged over 5 runs) across 60 games and number of games where an offline agent trained using the DQN Replay Dataset achieves better scores than a fully-trained DQN~(Nature) agent.
    Offline DQN~(Adam) significantly outperforms DQN~(Nature) and needs further investigation.
    Offline REM surpasses the gains~(119\%) from fully-trained C51, a strong online agent.
    }\label{table:results_summary}
      \centering
      \small
    \begin{tabular}{llrr}
    \toprule
    Offline agent & Median & $>\,$DQN \\
    \midrule
    DQN~(Nature) & 83.4\%  & 17 \\
    DQN~(Adam) & 111.9\% & 41 \\
    Ensemble-DQN & 111.0\% & 39\\
    Averaged Ensemble-DQN & 112.1\% & 43 \\
    QR-DQN     & 118.9\% & 45 \\
    REM      & \textbf{123.8}\%  & \textbf{49} \\
    \bottomrule
    \end{tabular}
\vspace{-0.55cm}
\end{table}

{\bf Comparison with QR-DQN.} QR-DQN modifies the DQN~(Nature) architecture to output $K$ values for each action using a multi-head $Q$-network and replaces RMSProp~\citep{tieleman2012lecture} with Adam~\citep{kingma2014adam} for optimization. To ensure a fair comparison with QR-DQN, we use the same multi-head $Q$-network as QR-DQN with $K=200$ heads~(Figure~\ref{fig:architechture}), where each head represents a $Q$-value estimate for REM and Ensemble-DQN. We also use Adam for optimization.

{\bf Additional Baselines.} To isolate the gains due to Adam in QR-DQN and our proposed variants,
we compare against a DQN baseline which uses Adam. We also evaluate Averaged Ensemble-DQN, a variant of Ensemble-DQN proposed by \citet{anschel2017averaged}, which uses the average of the predicted target $Q$-values as the Bellman target for training each parameterized $Q$-function.
This baseline determines whether the random combinations of REM provide any significant benefit over simply using an ensemble of
predictors to stabilize the Bellman target.

{\bf Results}. The main comparisons are presented in \figref{fig:batch_median}.
\tabref{table:results_summary} shows the comparison of baselines with REM and Ensemble-DQN. Surprisingly, DQN with Adam bridges the gap in asymptotic performance between QR-DQN and DQN~(Nature) in the offline setting. Offline Ensemble-DQN does not improve upon this strong DQN baseline showing that its naive ensembling approach is inadequate. Furthermore, Averaged Ensemble-DQN performs only slightly better than Ensemble-DQN. In contrast, REM exploits offline data more effectively than other agents, including QR-DQN, when trained for more gradient updates. Learning curves of all offline agents can be found in \figref{fig:batch_agents}.

{\bf Hypothesis about effectiveness of REM}. The gains from REM over Averaged Ensemble-DQN suggest that the effectiveness of REM is due to the noise from randomly ensembling Q-value estimates leading to more robust training, analogous to dropout. Consistent with this hypothesis, we also find that offline REM with separate $Q$-networks~(with more variation in individual $Q$-estimates) performs better asymptotically and learns faster than a multi-head $Q$-network~(Figure~\ref{fig:multi_head_vs_multi_network_rem}).
Furthermore, randomized ensembles as compared to a minimum over the $Q$-values for the target perform substantially better in our preliminary experiments on 5 games and requires further investigation.

\begin{figure}[t]
    \centering
    \includegraphics[width=0.95\linewidth]{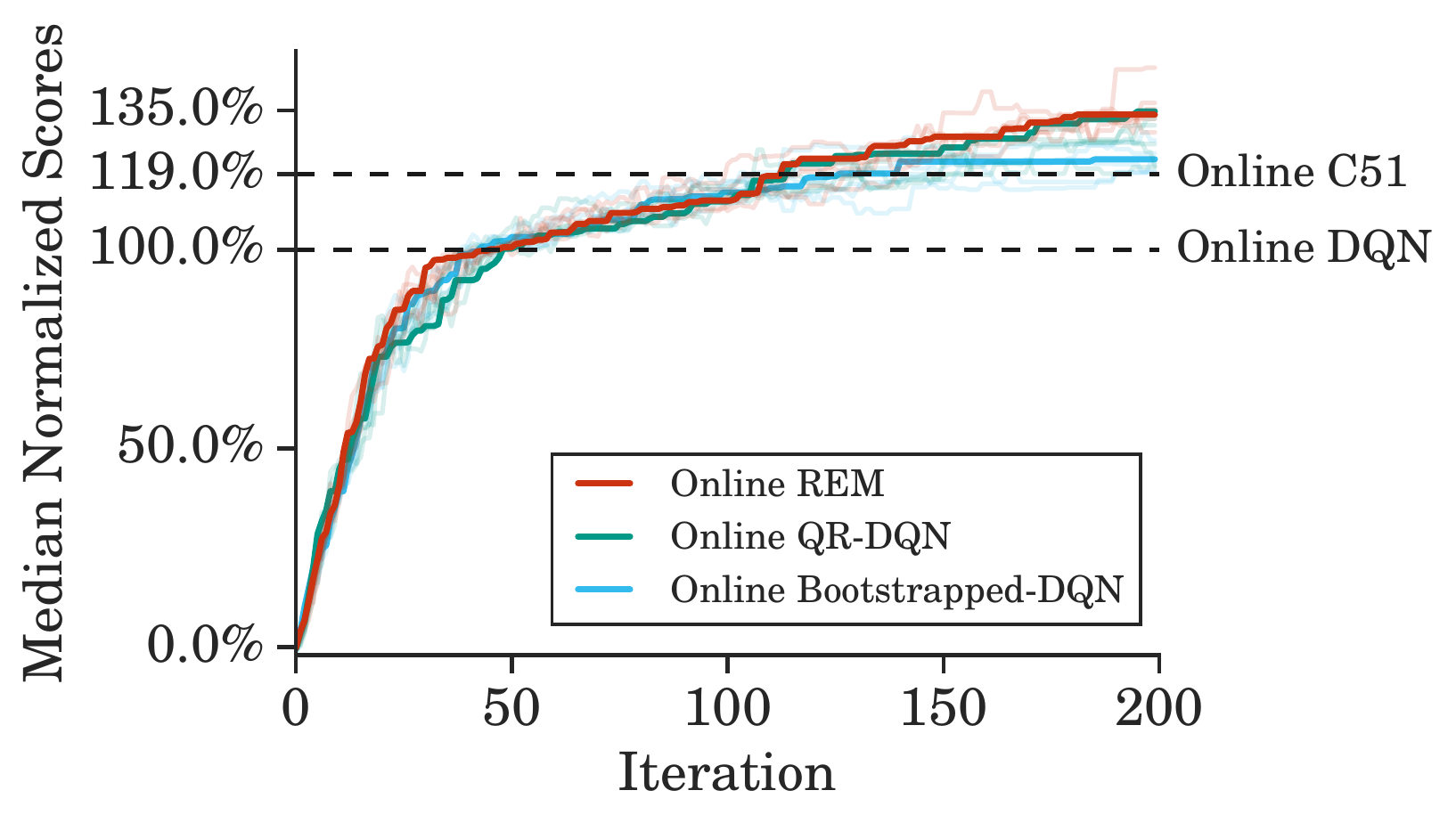}
    \vspace{-0.2cm}
    \caption{\textbf{Online REM \versus baselines}. Median normalized evaluation scores averaged over 5 runs~(shown as traces) across stochastic version of 60 Atari 2600 games of online agents trained for 200 million frames~(standard protocol). Online REM with 4 $Q$-networks performs comparably to online QR-DQN. Please refer to Figure~\ref{fig:online_agents} for learning curves.}\label{fig:online_rem}
    \vspace{-0.4cm}
\end{figure}

\begin{figure*}[t]
\begin{center}
\includegraphics[width=\linewidth]{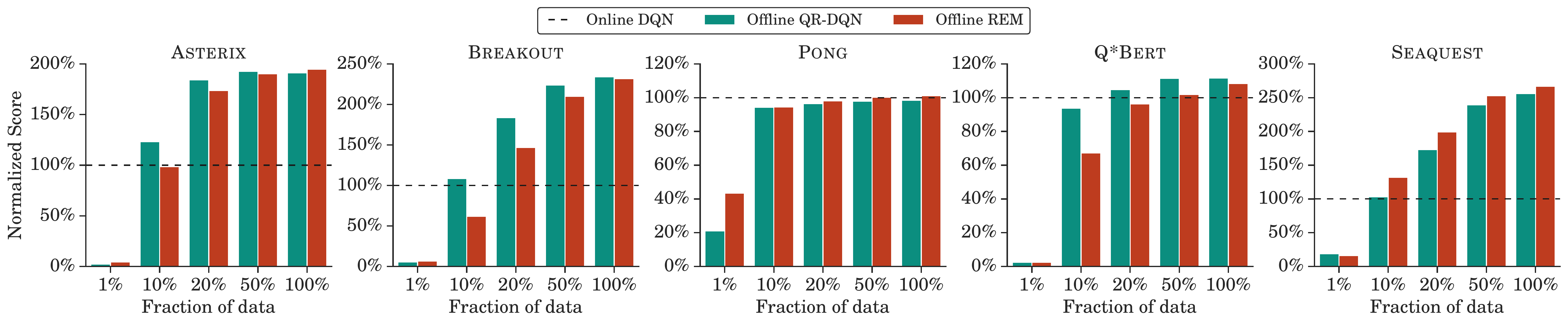}
\end{center}
\vspace{-0.2cm}
\caption{{\bf Effect of Offline Dataset Size.} Normalized scores~(averaged over 5 runs) of QR-DQN and multi-head REM trained offline on stochastic version of 5 Atari 2600 games for 5X gradient steps using only a fraction of the entire DQN Replay Dataset~(200M frames) obtained via randomly subsampling trajectories. With only 10\% of the entire replay dataset, REM and QR-DQN approximately recover the performance of fully-trained DQN.}\label{fig:random_subsampling_plots}
\vspace{-0.1cm}
\end{figure*}

\begin{figure*}[t]
\begin{center}
\includegraphics[width=\linewidth]{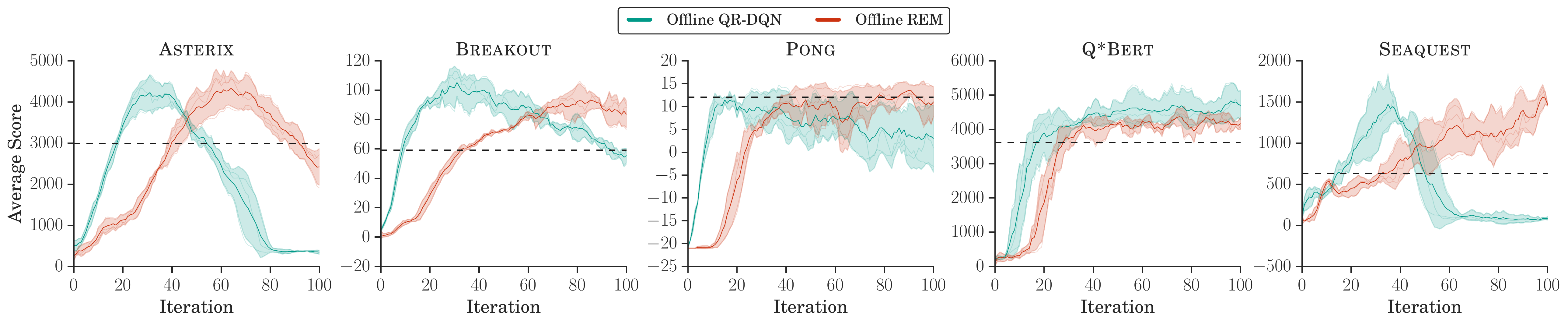}
\end{center}
\vspace{-0.25cm}
\caption{{\bf Offline RL with Lower Quality Dataset.} Normalized scores~(averaged over 3 runs) of QR-DQN and multi-head REM trained offline on stochastic version of 5 Atari 2600 games for 5X gradient steps using logged data from online DQN trained only for 20M frames~(20 iterations). The horizontal line shows the performance of best policy found during DQN training for 20M frames which is significantly worse than fully-trained DQN. We observe qualitatively similar results to
offline setting with entire replay dataset.}\label{fig:random_data_offline}
\vspace{-0.3cm}
\end{figure*}

\subsection{Does REM work in the online setting?}
In online RL, learning and data generation are tightly coupled, \ie\ an
agent that learns faster also collects more relevant data.
We ran online REM with $4$ separate $Q$-networks because of the better convergence speed
over multi-head REM in the offline setting. For data collection, we
use $\epsilon$-greedy with a randomly sampled $Q$-estimate
from the simplex for each episode, similar to Bootstrapped DQN.
We follow the standard online RL protocol on Atari and use a fixed
replay buffer of 1M frames. 

To estimate the gains from the REM objective~\eqref{eq:sqn} in the online
setting, we also evaluate Bootstrapped-DQN with identical
modifications~(\eg separate $Q$-networks) as online
REM. Figure~\ref{fig:online_rem} 
shows that REM performs on par with
QR-DQN and considerably outperforms Bootstrapped-DQN.
This shows that we can use the insights gained from the offline setting with appropriate
design choices~(\eg~exploration, replay buffer) to create effective online methods.

\vspace{-0.1cm}
\section{Important Factors in Offline RL}
\label{sec:offline_factors}
\begin{figure*}[t]
  \begin{center}
      \includegraphics[width=0.77\linewidth]{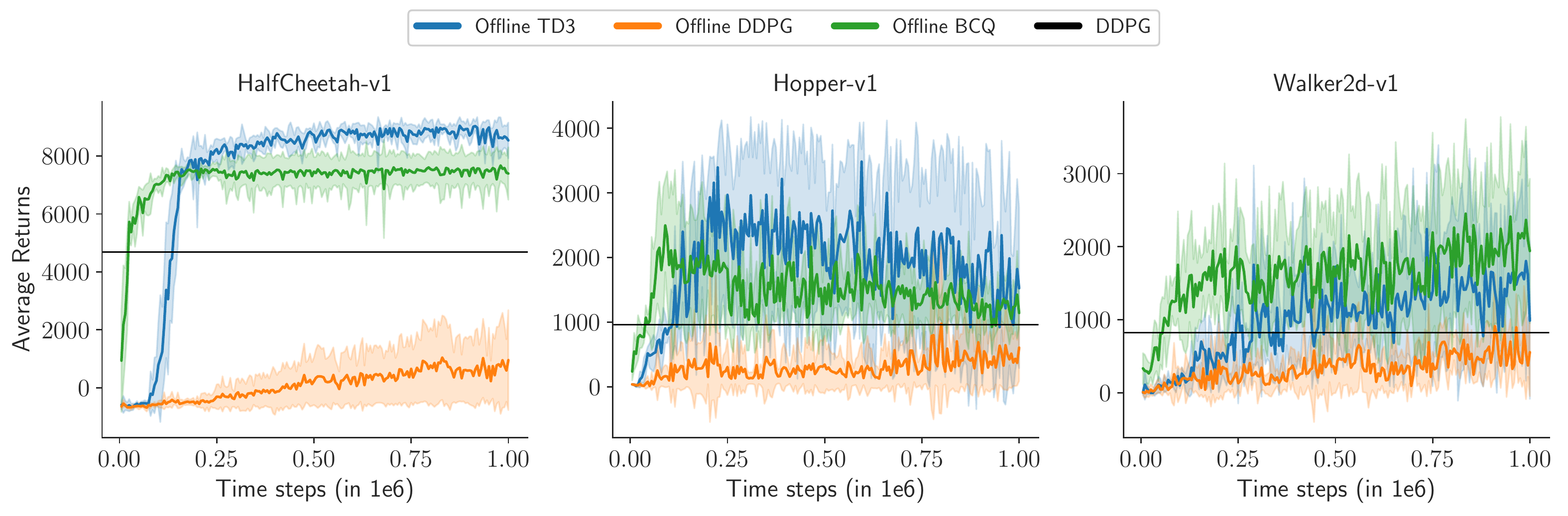}
     \vspace{-0.1cm}
    \caption{\textbf{Offline Continuous Control Experiments}. We examine the performance of DDPG, TD3 and BCQ~\citep{fujimoto2018off} trained using identical offline data on three standard MuJoCo environments. We plot the mean performance~(evaluated without exploration noise) across 5 runs. The shaded area represents a standard deviation. The bold black line measures the average return of episodes
      contained in the offline data collected using the DDPG agent (with exploration noise). 
      Similar to our results on Atari 2600 games, we find that recent off-policy algorithms perform quite well in the offline setting with entire DDPG replay data.}\label{fig:cc_results}
  \end{center}
\vspace{-0.5cm}
\end{figure*}

{\bf Dataset Size and Diversity}. Our offline learning results indicate that 50 million tuples per game from DQN~(Nature)
are sufficient to obtain good online performance on most of the Atari 2600 games. Behavior cloning performs poorly on the DQN Replay Dataset due to its diverse composition. We hypothesize that the size of the DQN Replay Dataset and its diversity~\citep{de2015importance} play a key role in the success of standard off-policy RL algorithms trained offline. 

To study the role of the offline dataset size, we
perform an ablation experiment with variable replay size.
We train offline QR-DQN and REM with reduced data obtained via randomly subsampling entire trajectories from the logged DQN
experiences, thereby maintaining the same data distribution.
Figure~\ref{fig:random_subsampling_plots} presents the performance of the offline REM and QR-DQN
agents with $N\%$ of the tuples in the DQN replay dataset where $N \in \{1, 10, 20, 50, 100\}$.
As expected, performance tends to increase as the fraction of data increases.
With $N \ge 10\%$, REM and QR-DQN still perform comparably to online DQN
on most of these games. However, the performance deteriorates drastically for $N=1\%$. 

To see the effect of quality of offline dataset, we perform another ablation where we train offline agents on the first $20$ million frames in the DQN Replay Dataset -- a lower quality dataset which roughly approximates exploration data with suboptimal returns. Similar to the offline results with the entire dataset, on most Atari games, offline REM and QR-DQN outperform the best policy in this dataset~(Figure~\ref{fig:random_data_offline} and Figure~\ref{fig:random_data_offline_all}), indicating that standard RL agents work well with sufficiently diverse offline datasets. 

{\bf Algorithm Choice}. Even though continuous control is not the focus of this paper, we reconcile the discrepancy between our findings~(Section~\ref{sec:part1_experiments}) and the claims of \citet{fujimoto2018off} that standard off-policy methods fail in the offline continuous control setting, even with large and diverse replay datasets. The results of \citet{fujimoto2018off} are based on the evaluation of a standard continuous control agent, called DDPG~\citep{lillicrap2015continuous}, and other more recent continuous control algorithms such as
TD3~\citep{fujimoto2018addressing} and SAC~\citep{haarnoja2018soft} are not considered in their study.

Motivated by the so-called \textit{final buffer} setting in \citet{fujimoto2018off}~(Appendix~\ref{sec:offline_cc}), we train a DDPG agent on continuous control MuJoCo
tasks~\citep{todorov2012mujoco} for 1 million time steps and store all of the experienced transitions. Using this dataset, we train standard off-policy agents including TD3 and DDPG completely offline. Consistent with our offline results on Atari games, offline TD3 significantly outperforms the data collecting DDPG agent and offline DDPG~(Figure~\ref{fig:cc_results}). Offline TD3 also performs comparably to BCQ~\citep{fujimoto2018off}, an algorithm designed specifically to learn from offline data.

\vspace{-0.15cm}
\section{Related work}
Our work is mainly related to batch RL~\footnote{To avoid confusion
with batch \versus minibatch optimization, we refer to batch RL as
offline RL in this paper.}~\citep{lange2012batch}.
Similar to \citep{ernst2005tree, riedmiller2005neural, jaques2019way}, we investigate batch off-policy RL,
which requires learning a good policy given a fixed dataset of interactions.
In our offline setup, we only assume access to samples from the behavior policy and focus
on $Q$-learning methods without any form of importance correction, as opposed to
\citep{swaminathanjoachims15c, liu2019off}.

Recent work~\citep{fujimoto2018off, kumar2019stabilizing,
wu2019behavior, siegel2020keep} reports that standard off-policy
methods trained on fixed datasets fail on continuous control
environments. \citet{fujimoto2019benchmarking} also observe that
standard RL algorithms fail on Atari 2600 games when trained offline using
trajectories collected by a single partially-trained DQN policy. The
majority of recent papers focus on the offline RL setting with
dataset(s) collected using a single data collection policy~(\eg random,
expert, \etc) and propose remedies by regularizing the learned policy
to stay close to training trajectories. These approaches improve
stability in the offline setting, however, they introduce additional
regularization terms and hyper-parameters, the selection of which is
not straightforward~\citep{wu2019behavior}. REM~(Section~\ref{sec:rem}) is
orthogonal to these approaches and can be easily combined with them.



This paper focuses on the offline RL setting on Atari 2600 games with
data collected from a large mixture of policies seen during the
optimization of a DQN agent, rather than a single Markovian behavior
policy. Our results
demonstrate that recent
off-policy deep RL algorithms~(\eg~TD3~\citep{fujimoto2018addressing},
QR-DQN~\citep{dabney2018distributional}) are effective in the offline
setting with sufficiently large and diverse offline datasets, without
explicitly correcting for the distribution
mismatch between the learned policy and the offline
dataset. 
We suspect that the suboptimal
performance of offline DQN~(Nature) is related to the
notion of off-policyness of large buffers established
by \citet{Zhang2017ADL}. However, robust deep $Q$-learning algorithms
such as REM are able to effectively exploit DQN Replay, given
sufficient number of gradient updates. 
Section~\ref{sec:offline_factors} shows that dataset size and its diversity, as well as choice of the RL algorithms significantly affect offline RL results and explains the discrepancy with recent work. 
Inspired by our work, \citet{cabi2019framework}
successfully apply distributional RL algorithms on large-scale offline robotic datasets. 

\vspace{-0.15cm}
\section{Future Work}

Since the \textit{(observation, action, next observation, reward)} tuples in DQN Replay Dataset are stored in the order they were experienced by online DQN during training, various data collection strategies for \textit{benchmarking} offline RL can be induced by subsampling the
replay dataset containing 200 million frames. For example, the first $k$ million frames from the DQN Replay Dataset emulate exploration data with suboptimal returns~(\eg Figure~\ref{fig:random_data_offline}) while the last $k$ million frames are analogous to near-expert data with stochasticity. Another option is to randomly subsample the entire dataset to create smaller offline datasets~(\eg Figure~\ref{fig:random_subsampling_plots}).
Based on the popularity and ease of experimentation on Atari 2600 games, the DQN Replay Dataset can be used for benchmarking offline RL in addition to continuous control setups such as BRAC~\citep{wu2019behavior}.

As REM simply uses a randomized $Q$-ensemble, it is straightforward to combine REM with existing $Q$-learning methods including distributional RL. REM can be used for improving value baseline estimation in policy gradient and actor-critic methods~\citep{weng2018PG}.
Using entropy regularization to create correspondence between $Q$-values and policies~(\eg\ SAC~\citep{haarnoja2018soft}) and applying REM to $Q$-values is another possibility for future work. REM can also be combined with behavior regularization methods
such as SPIBB~\citep{laroche2017safe} and BCQ~\citep{fujimoto2018off} to create better offline RL algorithms.

Our results also emphasize the need for a rigorous characterization of the role of  \textit{generalization} due to deep neural nets when learning from offline data collected from a large mixture of (diverse) policies. We also leave further investigation of the exploitation ability of distributional RL as well as REM to future work. 

Analogous to supervised learning, the offline agents
exhibit overfitting, \ie after certain number of gradient updates, their performance starts to deteriorates. To avoid such overfitting, we currently employ online policy evaluation for early stopping, however, ``true'' offline RL requires offline policy evaluation for hyperparameter tuning and early stopping. We also observe divergence \wrt Huber loss in our reduced data experiments with $N=1\%$. This suggests the need for offline RL methods that maintain reasonable performance throughout learning~\citep{garcia2015comprehensive}.


Experience replay-based algorithms can be more sample efficient than model-based approaches~\citep{van2019use}, and using the DQN Replay Dataset on Atari 2600 games for designing non-parametric replay models~\citep{pan2018organizing} and parametric world models~\citep{kaiser2019model} is another promising direction for improving sample-efficiency in RL. 



\vspace{-0.15cm}
\section{Conclusion}
This paper studies offline RL on Atari 2600 games based on
logged experiences of a DQN agent. The paper demonstrates that standard RL methods can learn to play Atari games from DQN Replay Dataset, better than the best behavior in the dataset. This is in contrast with existing work, which claims standard methods fail in the offline setting. The DQN Replay Dataset can serve as a benchmark
for offline RL. 
These results present a positive view that robust RL
algorithms can be developed which can effectively learn from large-scale offline datasets. 
REM strengthens this optimistic perspective by showing that even simple ensemble methods can 
be effective in the offline setting.

Overall, the paper indicates the potential of offline RL for creating a \textit{data-driven} RL paradigm where one could pretrain RL agents with large amounts of existing diverse datasets before further collecting new data via exploration, thus creating \textit{sample-efficient} agents that can be deployed and continually learn in the real world.

\vspace{-0.15cm}
\section*{Acknowledgements}
We thank Pablo Samuel Castro for help in understanding and debugging issues with
the Dopamine codebase and reviewing an early draft of the paper. We thank Andrew Helton, 
Christian Howard, Robert Dadashi, Marlos C. Machado, Carles Gelada, Dibya Ghosh and Liam Fedus for helpful discussions. 
We also acknowledge
Marc G. Bellemare, Zafarali Ahmed, Ofir Nachum, George Tucker, Archit Sharma, Aviral Kumar,
William Chan and anonymous ICML reviewers for their review of the paper.

\vspace{-0.15cm}
{\small \bibliography{main}}
\bibliographystyle{icml2020}

\appendix
\cleardoublepage
\icmltitlerunning{Supplementary Material for \title}
\counterwithin{figure}{section}
\section{Appendix}

\subsection{Proofs}

\textbf{Proposition 1}.
{\it
Consider the assumptions:
(a) The distribution $\ermP_{\Delta}$ has full support over the entire $(K-1)$-simplex.
(b) Only a finite number of distinct $Q$-functions globally minimize the loss in \eqref{eq:dqn_loss}.
(c) $Q^*$ is defined in terms of the MDP induced by the data distribution $\mathcal{D}$.
(d) $Q^*$ lies in the family of our function approximation.
Then at the global minimum of $\calL({\theta})$~\eqref{eq:sqn} for multi-head Q-network :
\setlist{nolistsep}
\begin{enumerate}[label=(\roman*)]
\itemsep0.15em
\item Under assumptions (a) and (b), all the $Q$-heads represent identical $Q$-functions.
\item Under assumptions (a)--(d), the common convergence point is $Q^*$.
\end{enumerate}
}

\textbf{Proof}. Part \textit{(i)}:
Under assumptions (a) and (b), we would prove by contradiction
that each $Q$-head should be identical to minimize the REM loss
$\calL({\theta})$~\eqref{eq:sqn}.
Note that we consider two $Q$-functions to be distinct only if they differ on any
state $s$ in $\setD$.

The REM loss $\calL({\theta}) = \expected_{\alpha \sim P^{\Delta}} \left[\calL(\alpha, \theta)\right]$
where $\calL(\alpha, \theta)$ is given by
\begin{align}\label{eq:loss_alpha_theta}
\calL(\alpha, \theta) ={}& \expected_{s, a, r, s' \sim \setD} \left[ \ld \big( \Delta^{\alpha}_{\theta}(s, a, r, s')
 \big) \right], \\
\Delta^{\alpha}_{\theta} =& \sum_{k} \alpha_{k} \Qt^{k}(s, a) - r - \gamma\max_{a'} \sum_{k} \alpha_{k}\Qtp^{k}(s', a') \nonumber
\end{align}
If the heads $\Qt^{i}$ and $\Qt^{j}$ don't converge to identical $Q$-values at the global minimum
of $\calL({\theta})$, it can be deduced using Lemma 1 that all the $Q$-functions given by the convex combination
$\alpha_i \Qt^{i} + \alpha_j \Qt^{j}$ such that $\alpha_i + \alpha_j = 1$ minimizes the loss in \eqref{eq:dqn_loss}.
This contradicts the assumption that only a finite number of distinct $Q$-functions globally minimize the
loss in \eqref{eq:dqn_loss}. Hence, all $Q$-heads represent an identical $Q$-function at the
global minimum of $\calL({\theta})$.

\textbf{Lemma 1}. {\it Assuming that the distribution $\ermP_{\Delta}$ has full support
over the entire $(K-1)$-simplex $\Delta^{\ermK - 1}$, then
at any global minimum of $\calL({\theta})$, the $Q$-function heads $\Qt^{k}$ for $k=1,\dots,K$ minimize
$\calL(\alpha, \theta)$ for any $\alpha \in \Delta^{\ermK - 1}$.}

\textbf{Proof}. Let $Q_{\alpha^{*}, \theta^{*}} =\sum_{k=1}^{K} \alpha^{*}_{k} Q^{k}_{\theta^{*}}(s, a)$
corresponding to the convex combination $\alpha^{*} = (\alpha^{*}_1, \cdots, \alpha^{*}_K)$
represents one of the global minima of
$\calL(\alpha, \theta)$~\eqref{eq:loss_alpha_theta} \ie\
$\calL(\alpha^{*}, \theta^{*}) = \underset{\alpha, \theta}{\text{min}}\ \calL(\alpha, \theta)$ where
$\alpha \in \Delta^{\ermK - 1}$.
Any global minima of $\calL(\theta)$ attains a value of $\calL(\alpha^{*}, \theta^{*})$ or higher since,
\begin{align}
\calL(\theta) ={}& \expected_{\alpha \sim P^{\Delta}} \left[\calL(\alpha, \theta)\right]\\ \nonumber
\ge&~ \expected_{\alpha \sim P^{\Delta}} \left[\calL(\alpha^{*}, \theta^{*})\right]
\ge \calL(\alpha^{*}, \theta^{*}) \nonumber
\end{align}

Let $Q^{k}_{\theta^{*}}(s,a) = w^{k}_{\theta^{*}} \cdot f_{\theta^{*}}(s,a)$ where
$f_{\theta^{*}}(s, a) \in \Real^{D}$ represent the shared features among the $Q$-heads
and $w^{k}_{\theta^{*}} \in \Real^{D}$ represent the weight vector in
the final layer corresponding to the $k$-th head. Note that $Q_{\alpha^{*}, \theta^{*}}$ can also
be represented by each of the individual $Q$-heads using a weight vector given by
convex combination $\alpha^{*}$ of weight vectors $(w^{1}_{\theta^{*}}, \cdots, w^{K}_{\theta^{*}})$, \textit{i.e.},
$Q(s, a) = \left(\sum_{k=1}^{K} \alpha^{*}_{k} w^{k}_{\theta^{*}}\right) \cdot f_{\theta^{*}}(s,a)$.

Let $\theta^{I}$ be such that
$Q^{k}_{\theta^{I}} = Q_{\alpha^{*}, \theta^{*}}$ for all $Q$-heads.
By definition of $Q_{\alpha^{*}, \theta^{*}}$, for all $\alpha \sim \ermP_{\Delta}$,
$\calL(\alpha, \theta^{I}) = \calL(\alpha^{*}, \theta^{*})$
which implies that $\calL(\theta^{I}) = \calL(\alpha^{*}, \theta^{*})$.
Hence, $\theta^{I}$ corresponds to one of the global minima of $\calL(\theta)$ and
any global minima of $\calL(\theta)$ attains a value of $\calL(\alpha^{*}, \theta^{*})$.

Since $\calL(\alpha, \theta) \ge \calL(\alpha^{*}, \theta^{*})$ for any $\alpha \in \Delta^{\ermK - 1}$,
for any $\theta^{M}$ such that $\calL(\theta^{M}) = \calL(\alpha^{*}, \theta^{*})$ implies
that $\calL(\alpha, \theta^{M}) = \calL(\alpha^{*}, \theta^{*})$ for any $\alpha \sim \ermP_{\Delta}$. Therefore,
at any global minimum of $\calL({\theta})$, the $Q$-function heads $\Qt^{k}$ for $k=1,\dots,K$ minimize
$\calL(\alpha, \theta)$ for any $\alpha \in \Delta^{\ermK - 1}$.

\subsection{Offline continuous control experiments}\label{sec:offline_cc}
We replicated the \textit{final buffer} setup as described by \citet{fujimoto2018off}: We train a DDPG~\citep{lillicrap2015continuous}
agent for 1 million time steps three standard MuJoCo continuous control environments in OpenAI gym~\citep{todorov2012mujoco, brockman2016openai}, adding $\mathcal{N}(0, 0.5)$
Gaussian noise to actions for high exploration, and store all experienced transitions. This collection procedure creates a dataset with a diverse set of
states and actions, with the aim of sufficient coverage. Similar to \citet{fujimoto2018off}, we train DDPG across 15 seeds,
and select the 5 top performing seeds for dataset collection.

Using this logged dataset, we train standard continuous control off-policy actor-critic methods namely DDPG
and TD3~\citep{fujimoto2018addressing} completely offline without any exploration. We also train a
Batch-Constrained deep $Q$-learning~(BCQ) agent, proposed by \citet{fujimoto2018off}, which restricts the action space
to force the offline agent towards behaving close to on-policy \wrt a subset of the given data.
We use the open source code generously provided by the authors at \url{https://github.com/sfujim/BCQ} and \url{https://github.com/sfujim/TD3}.
We use the hyperparameters mentioned in \citep{fujimoto2018addressing, fujimoto2018off} except offline TD3 which uses a learning rate of
0.0005 for both the actor and critic.

Figure~\ref{fig:cc_results} shows that offline TD3 significantly outperforms the behavior policy which collected the offline data as well as the
offline DDPG agent. Noticeably, offline TD3 also performs comparably to BCQ, an algorithm designed specifically to
learn from arbitrary, fixed offline data. While \citet{fujimoto2018off} attribute the failure to learn in the offline setting to
extrapolation error~(\ie~the mismatch between the offline dataset and true state-action visitation of the current policy),
our results suggest that failure to learn from diverse offline data 
may be linked to extrapolation error for only weak exploitation agents such as DDPG.

\subsection{Score Normalization}\label{sec:score_normalization}
The improvement in normalized performance of an offline agent,
expressed as a percentage, over an online DQN~(Nature)~\citep{mnih2015human} agent is calculated as: $100 \times (\mathrm{Score_{normalized}} - 1)$ where:
\begin{align}\label{eq:score_normalize}
\mathrm{Score_{normalized}} ={}& \frac{\mathrm{Score_{Agent}}\ -\ \mathrm{Score_{min}}}
{\mathrm{Score_{max}} - \mathrm{Score_{min}}},\\
\mathrm{Score_{min}} &= \min(\mathrm{Score_{DQN}},\ \mathrm{Score_{Random}}) \nonumber\\
\mathrm{Score_{max}} &= \max(\mathrm{Score_{DQN}},\ \mathrm{Score_{Random}}) \nonumber
\end{align}
Here, $\mathrm{Score_{DQN}}$, $\mathrm{Score_{Random}}$ and $\mathrm{Score_{Agent}}$ are the mean evaluation scores averaged over 5 runs.
We chose not to measure performance in terms of percentage of online DQN scores alone
because a tiny difference relative to the random agent on some
games can translate into hundreds of percent in DQN score difference. Additionally, the max is needed
since DQN performs worse than a random agent on the games Solaris and Skiing.

\subsection{Hyperparameters \& Experiment Details}\label{sec:hyperparameters}
In our experiments, we used the hyperparameters provided in Dopamine baselines~\citep{castro2018dopamine} and report them for
completeness and ease of reproducibility in \tabref{table:hyperparams}. As mentioned by Dopamine's GitHub repository,
changing these parameters can significantly affect performance, without necessarily being indicative of an algorithmic difference.
We will also open source our code to further aid in reproducing our results.

\begin{table*}[t]
\small
\caption{The hyperparameters used by the offline and online RL agents in our experiments.}
\centering
\begin{tabular}{lrr}
\toprule
Hyperparameter & \multicolumn{2}{r}{setting (for both variations)} \\
\midrule
Sticky actions && Yes        \\
Sticky action probability && 0.25\\
Grey-scaling && True \\
Observation down-sampling && (84, 84) \\
Frames stacked && 4 \\
Frame skip~(Action repetitions) && 4 \\
Reward clipping && [-1, 1] \\
Terminal condition && Game Over \\
Max frames per episode && 108K \\
Discount factor && 0.99 \\
Mini-batch size && 32 \\
Target network update period & \multicolumn{2}{r}{every 2000 updates} \\
Training steps per iteration && 250K \\
Update period every && 4 steps \\
Evaluation $\epsilon$ && 0.001 \\
Evaluation steps per iteration && 125K \\
$Q$-network: channels && 32, 64, 64 \\
$Q$-network: filter size && $8\times8$, $4\times4$, $3\times3$\\
$Q$-network: stride && 4, 2, 1\\
$Q$-network: hidden units && 512 \\
Multi-head $Q$-network: number of $Q$-heads && 200 \\
Hardware && Tesla P100 GPU \\
\midrule
Hyperparameter & Online & Offline\\
\midrule
Min replay size for sampling & 20,000 & - \\
Training $\epsilon$~(for $\epsilon$-greedy exploration) & 0.01 & - \\
$\epsilon$-decay schedule & 250K steps & - \\
Fixed Replay Memory & No & Yes \\
Replay Memory size & 1,000,000 steps & 50,000,000 steps \\
Replay Scheme & Uniform & Uniform \\
Training Iterations & 200 & 200 or 1000 \\
\bottomrule
\end{tabular}
\label{table:hyperparams}
\end{table*}

The Atari environments~\citep{bellemare2013arcade} used in our experiments are stochastic due to sticky actions~\citep{machado2018revisiting}, \ie\ there is 25\% chance at every time step that the environment will execute the agent's previous action again, instead of the agent's new action. All agents~(online or offline) are compared using the best evaluation score~(averaged over 5 runs) achieved during training where the evaluation is done online every training iteration using a $\epsilon$-greedy policy with $\epsilon=0.001$. We report offline training results with same hyperparameters over 5 random seeds of the DQN replay data collection, game simulator and network initialization.

{\bf DQN replay dataset collection}. For collecting the offline data used in our experiments, we use online DQN~(Nature)~\citep{mnih2015human} with the RMSprop~\citep{tieleman2012lecture} optimizer. The DQN replay dataset, $\setB_{\mathrm{DQN}}$, consists of approximately 50 million experience tuples for each run per game corresponds to 200 million frames due to frame skipping of four, \ie~repeating a selected action for four consecutive frames. Note that the total dataset size is approximately 15 billion tuples~(
$50~\frac{\mathrm{million\ tuples}}{\mathrm{agent}} * 5~\frac{\mathrm{agents}}{\mathrm{game}} * 60~\mathrm{games}$).

{\bf Optimizer related hyperparameters}. For existing off-policy agents, step size and optimizer were taken as published. Offline DQN~(Adam) and all the offline agents with multi-head $Q$-network~(Figure~\ref{fig:architechture}) use the Adam optimizer~\citep{kingma2014adam} with same hyperparameters as online QR-DQN~\citep{dabney2018distributional}~($\mathrm{lr}=0.00005$, $\epsilon_{\mathrm{Adam}}=0.01/32$).
Note that scaling the loss has the same effect as inversely scaling $\epsilon_{\mathrm{Adam}}$ when using Adam.

{\bf Online Agents}. For online REM shown in Figure~\ref{fig:batch_median}b, we performed hyper-parameter tuning over $\epsilon_{\mathrm{Adam}}$ in ($0.01/32$, $0.005/32$, $0.001/32$) over 5 training games~(Asterix, Breakout, Pong, Q*Bert, Seaquest) and evaluated on the full set of 60 Atari 2600 games using the best setting~($\mathrm{lr}=0.00005$, $\epsilon_{\mathrm{Adam}}=0.001/32$). Online REM uses 4 $Q$-value estimates calculated using separate $Q$-networks where each network has the same architecture as originally used by online DQN~(Nature). Similar to REM, our version of Bootstrapped-DQN also uses 4 separate $Q$-networks and Adam optimizer with identical hyperaparmeters~($\mathrm{lr}=0.00005$, $\epsilon_{\mathrm{Adam}}=0.001/32$).

{\bf Wall-clock time for offline experiments.} The offline experiments are approximately 3X faster than the online experiments for the same number of gradient steps on a P100 GPU. In Figure~\ref{fig:batch_median}, the offline agents are trained for 5X gradient steps, thus, the experiments are 1.67X slower than running online DQN for 200 million frames~(standard protocol). Furthermore, since the offline experiments do not require any data generation, using tricks from supervised learning such as using much larger batch sizes than 32 with TPUs / multiple GPUs would lead to a significant speed up.

\subsection*{Additional Plots \& Tables}

\begin{table*}[h]
    \caption{Median normalized scores~(Section~\ref{sec:score_normalization}) across stochastic version of 60 Atari 2600 games, measured as percentages
    and number of games where an agent achieves better scores than a fully trained online DQN~(Nature) agent. All the offline
    agents below are trained using the DQN replay dataset. The entries of the table without any suffix report training results
    with the five times as many gradient steps as online DQN while the entires with suffix~($1x$) indicates the same number of gradient
    steps as the online DQN agent. All the offline agents except DQN use the same multi-head architecture
    as QR-DQN.}\label{table:perc_scores}
    \centering
    \begin{minipage}{.48\linewidth}
      \centering
      \footnotesize
      \begin{tabular}{llrr}
      \toprule
      Offline agent & Median & $>\,$DQN \\
      \midrule
      DQN~(Nature)~($1x$)           &  74.4\%  & 10 \\
      DQN~(Adam)~($1x$)     & 104.6\% & 39 \\
      Ensemble-DQN~($1x$)  &  92.5\% & 26 \\
      Averaged Ensemble-DQN~($1x$) & 88.6\% & 24 \\
      QR-DQN~($1x$)        &  \textbf{115.0}\% & \textbf{44}  \\
      REM~($1x$)           &  103.7\% & 35  \\
      \bottomrule
      \end{tabular}
    \end{minipage}
    \begin{minipage}{.48\linewidth}
      \centering
      \footnotesize
      \begin{tabular}{llrr}
      \toprule
      Offline agent & Median & $>\,$DQN \\
      \midrule
      DQN~(Nature)      & 83.4\%  & 17 \\
      DQN~(Adam) & 111.9\% & 41 \\
      Ensemble-DQN & 111.0\% & 39\\
      Averaged Ensemble-DQN & 112.1\% & 43 \\
      QR-DQN     & 118.9\% & 45 \\
      REM      & \textbf{123.8}\%  & \textbf{49} \\
      \bottomrule
      \end{tabular}
    \end{minipage}
\vspace{-0.35cm}
\end{table*}

\begin{figure*}
  \begin{center}
    \footnotesize
    \begin{tabular}{@{}c@{}c@{}}
      \includegraphics[width=0.5\linewidth]{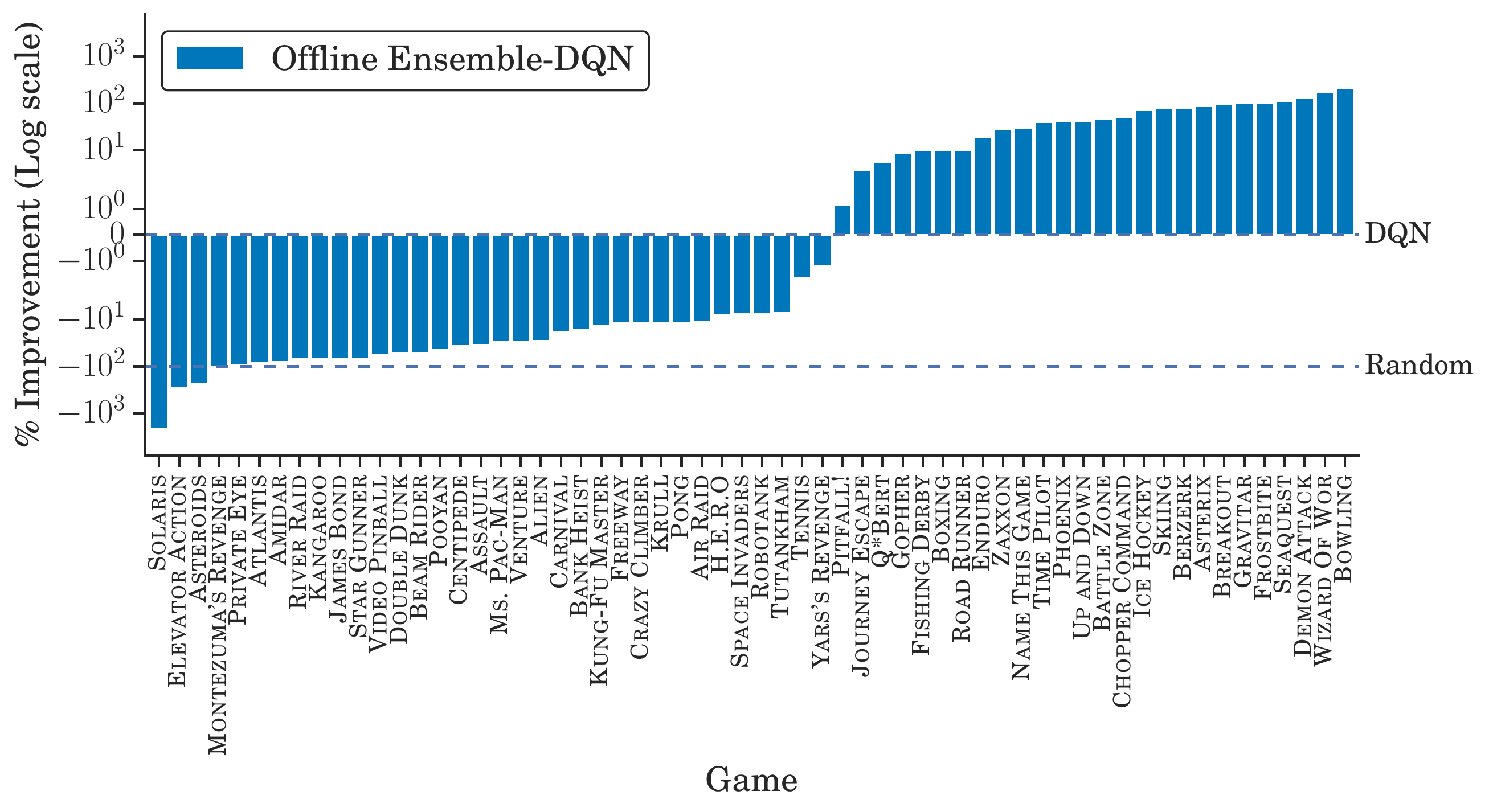} & \includegraphics[width=0.5\linewidth]{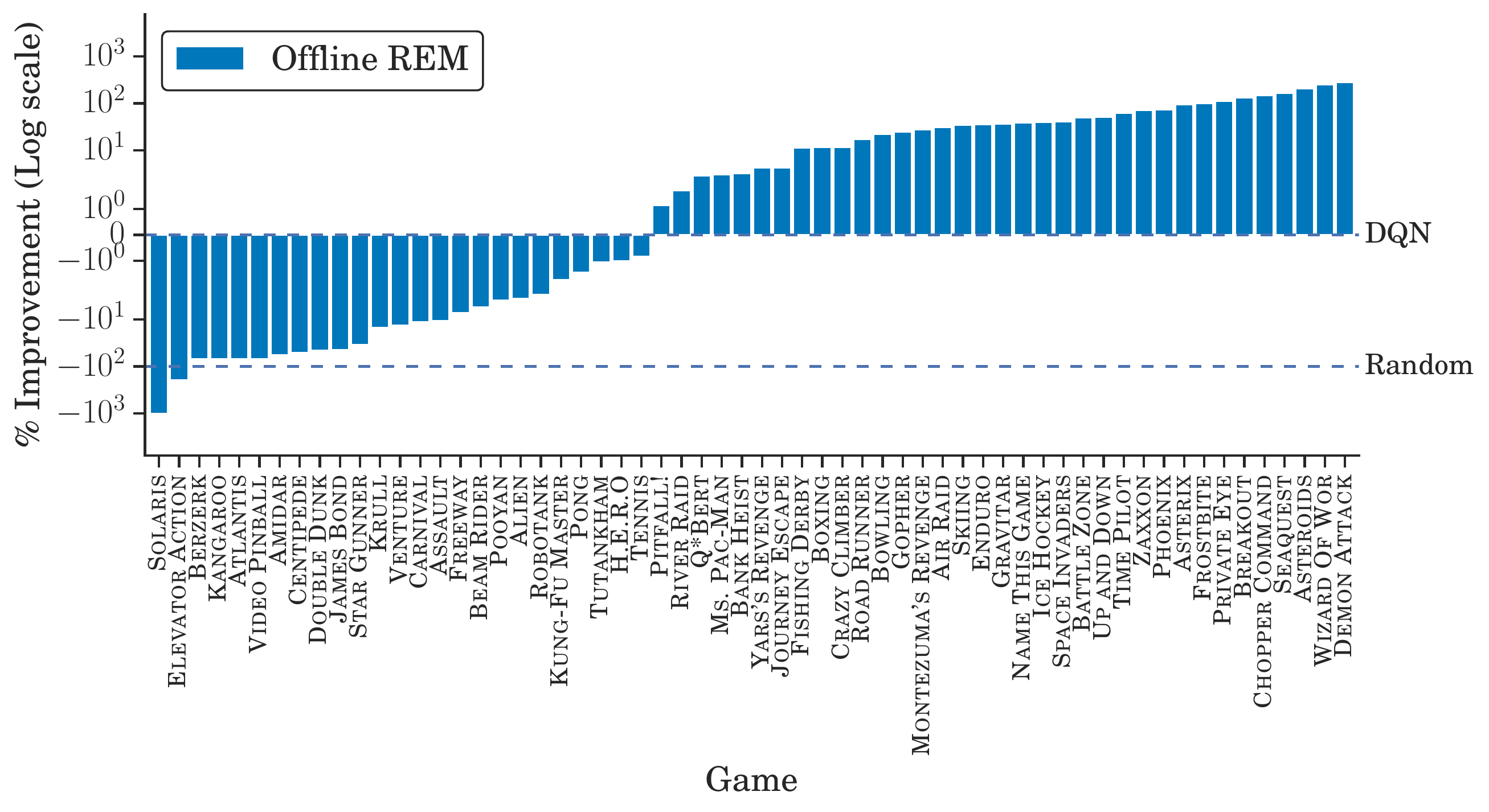} \\
      (a) & (b)
    \end{tabular}
    \caption{Normalized Performance improvement (in \%) over online DQN~(Nature), per game, of (a) offline Ensemble-DQN and (b) offline REM
    trained using the DQN replay dataset for same number of gradient steps as online DQN. The normalized online score for each game is 0.0 and 1.0 for the worse and better performing agent among fully trained online DQN and random agents respectively.} \label{fig:convex_dqn}
  \end{center}
\end{figure*}

\begin{figure*}[h]
  \begin{center}
    \footnotesize
    \begin{tabular}{@{}c@{}c@{}}
      \includegraphics[width=0.5\linewidth]{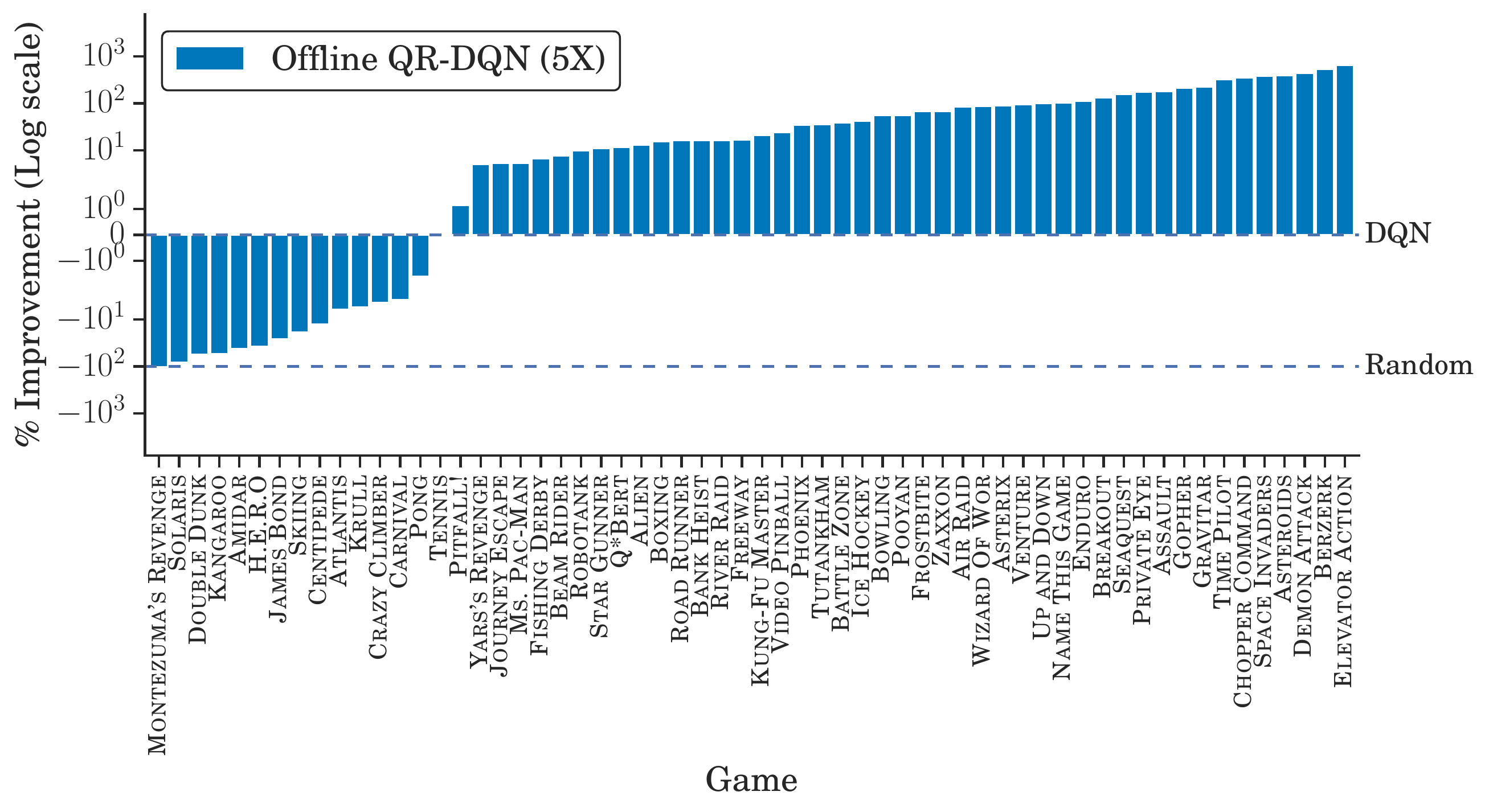} & \includegraphics[width=0.5\linewidth]{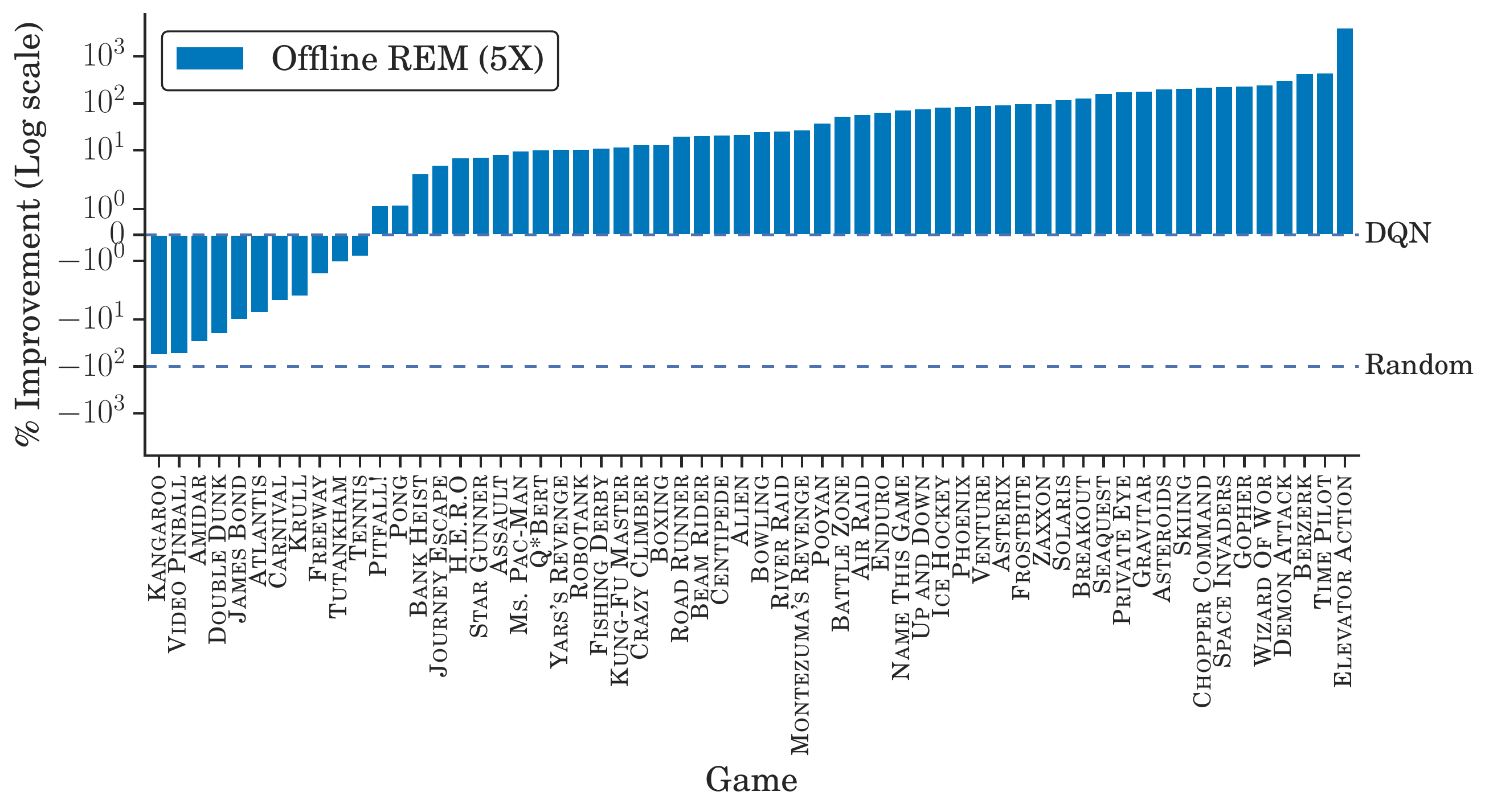} \\
      (a) & (b)
    \end{tabular}
    \caption{Normalized Performance improvement (in \%) over online DQN~(Nature), per game, of (a) offline QR-DQN~(5X) (b) offline REM~(5X)
    trained using the DQN replay dataset for five times as many gradient steps as online DQN. The normalized online score for each game is 0.0 and 1.0 for the worse and better performing agent among fully trained online DQN and random agents respectively.} \label{fig:batch_longer}
  \end{center}
\vspace{-0.4cm}
\end{figure*}

\begin{figure*}[h]
  \begin{center}
      \includegraphics[width=0.95\linewidth]{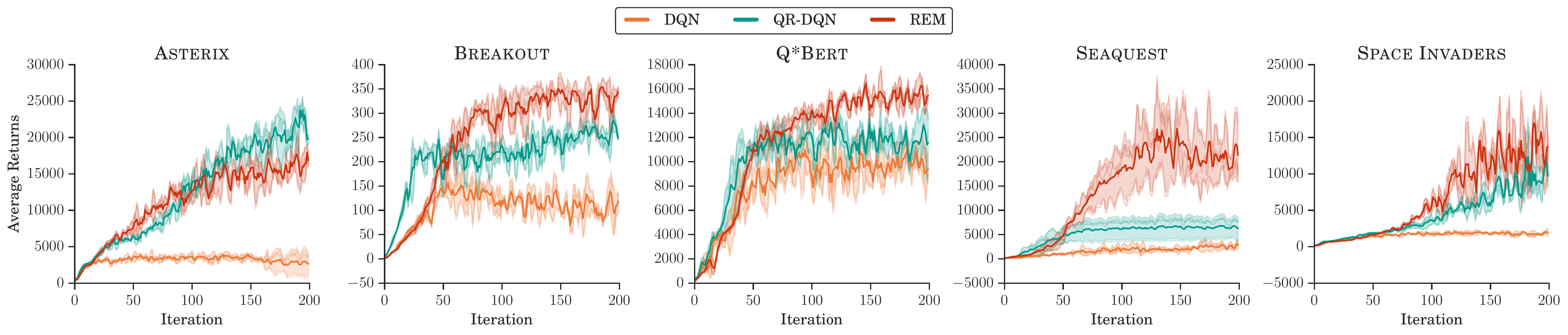}
    \caption*{{\bf Online REM \versus baselines}. Scores for online agents trained for 200 million ALE frames. Scores are averaged over 3 runs (shown as traces) and smoothed over a sliding window of 5 iterations and error bands show standard deviation.}\label{fig:online_rem_5_games}
  \end{center}
\end{figure*}

\begin{figure*}[t]
  \centering
  \footnotesize
  \begin{tabular}{@{}c@{}}
    \includegraphics[width=.75\linewidth]{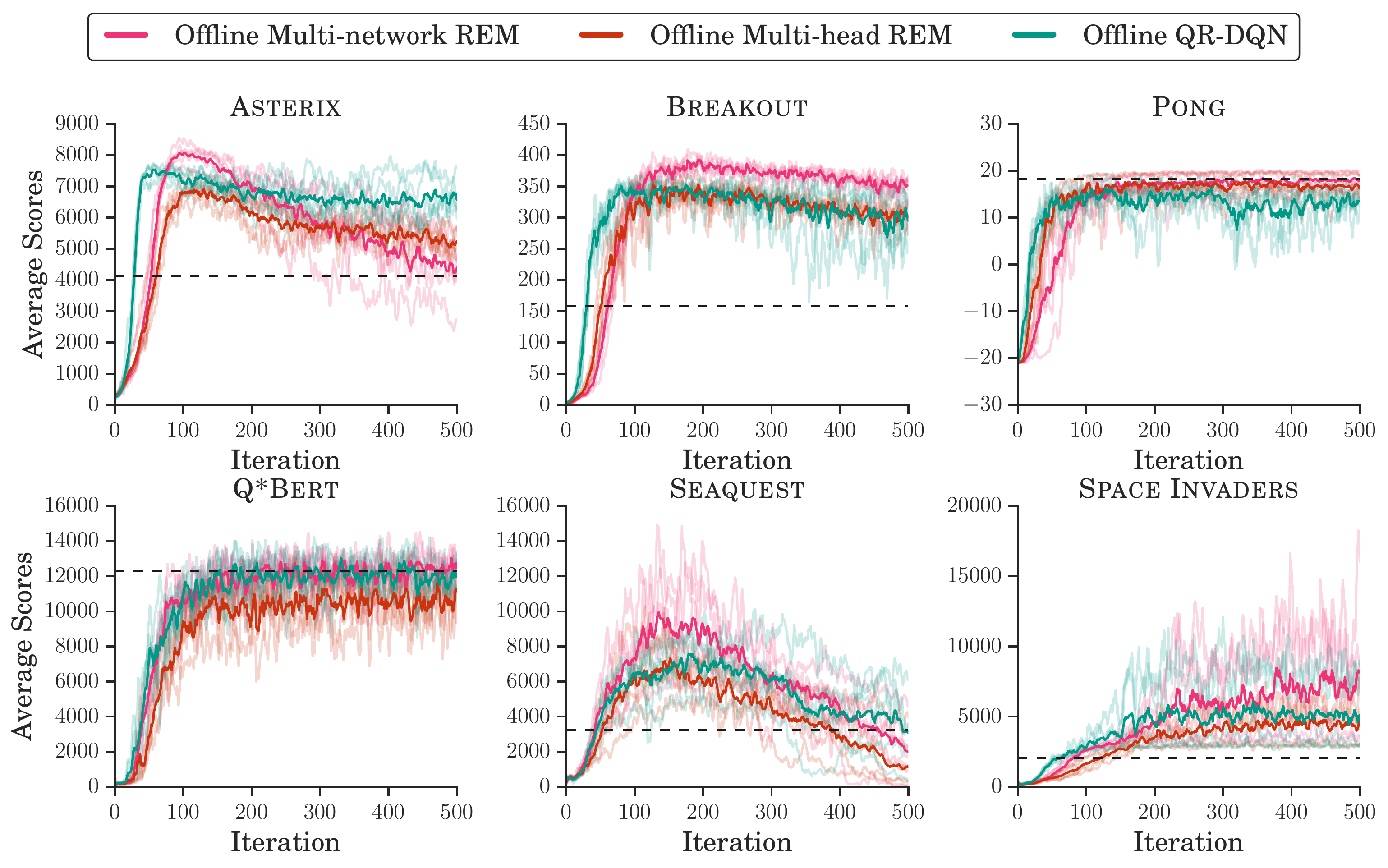} \\
    \small (a) REM with $4$ $Q$-value estimates~($K = 4$)
  \end{tabular}
  \vspace{0.1cm}\\
  \begin{tabular}{@{}c@{}}
    \includegraphics[width=.75\linewidth]{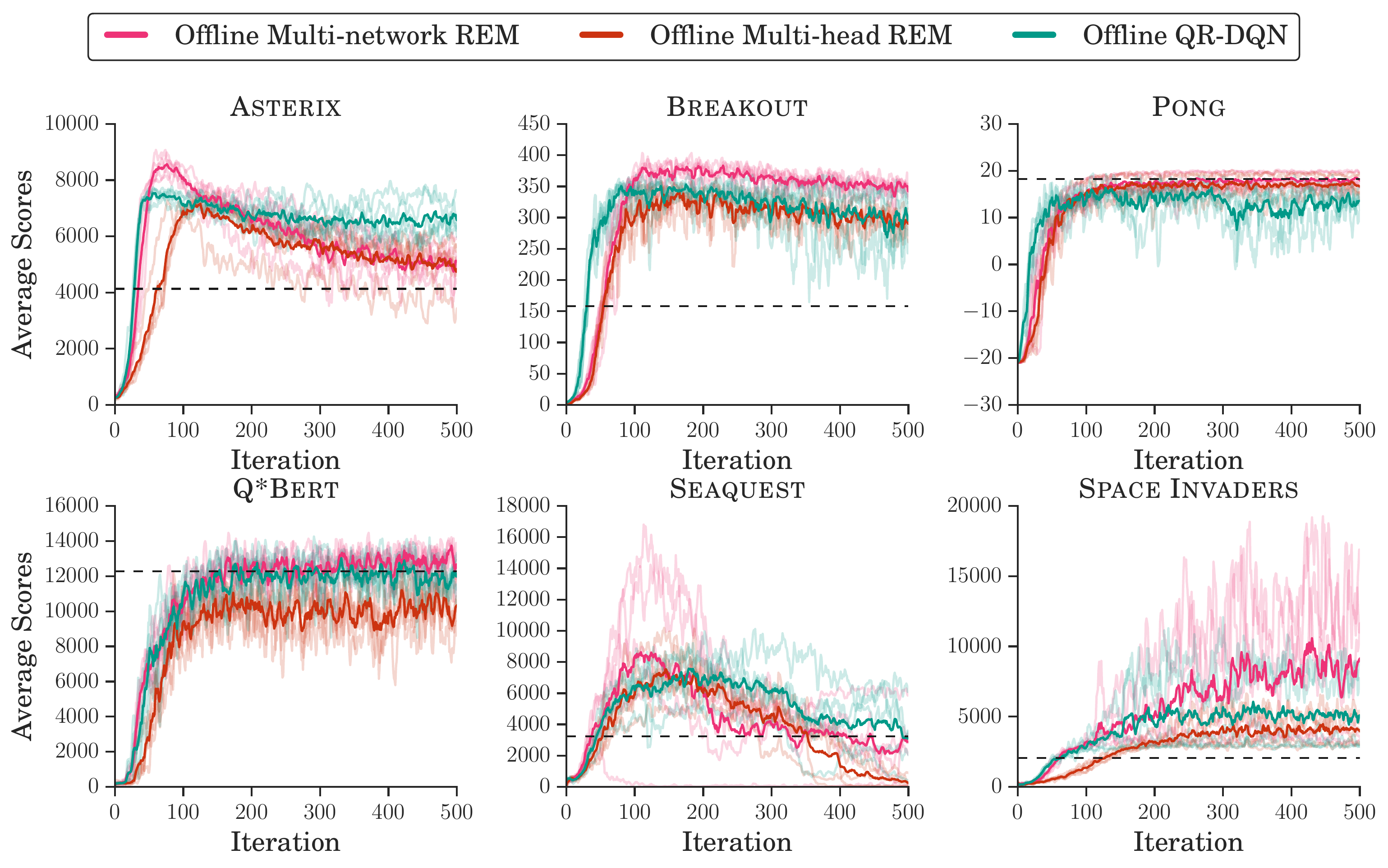} \\
    \small (b) REM with $16$ $Q$-value estimates~($K = 16$)
  \end{tabular}
    \caption{
    {\bf REM with Separate $Q$-networks.}
    Average online scores of offline REM variants with different architectures and QR-DQN trained on stochastic version of 6 Atari 2600 games for 500 iterations using
    the DQN replay dataset. The scores are averaged over 5 runs (shown as traces) and smoothed over a sliding window of 5 iterations and error bands show standard deviation.
    The multi-network REM and the multi-head REM employ $K$ $Q$-value estimates computed
    using separate $Q$-networks and $Q$-heads of a multi-head $Q$-network respectively and are optimized with
    identical hyperparameters. Multi-network REM improves upon the multi-head REM indicating that the more diverse $Q$-estimates
    provided by the separate $Q$-networks improve performance of REM over $Q$-estimates provided by the multi-head $Q$-network
    with shared features.}\label{fig:multi_head_vs_multi_network_rem}
\end{figure*}

\begin{figure*}[t]
\begin{center}
\includegraphics[width=0.90\linewidth]{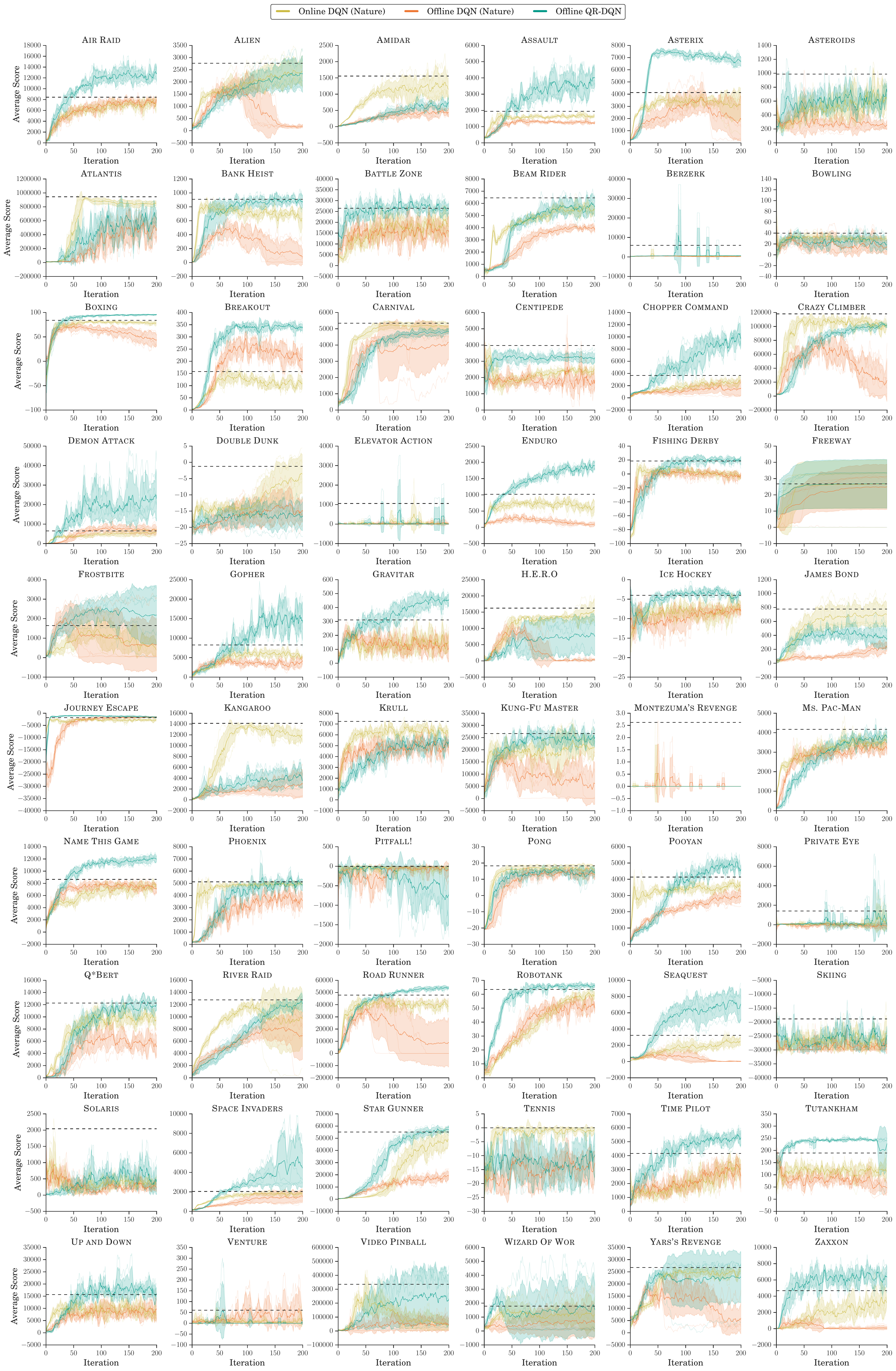}
\end{center}
\vspace{-0.3cm}
\caption{Average evaluation scores across stochastic version of 60 Atari 2600 games for online DQN, offline DQN and offline QR-DQN
trained for 200 iterations. The offline agents are trained using the DQN replay dataset. The scores are averaged over 5
runs~(shown as traces) and smoothed over a sliding window of 5 iterations and error bands show standard deviation. The horizontal line shows
the performance of the best policy~(averaged over 5 runs) found during training of online DQN.}
\label{fig:dqn_vs_batch}
\end{figure*}

\begin{figure*}[t]
\centering
\includegraphics[width=0.90\linewidth]{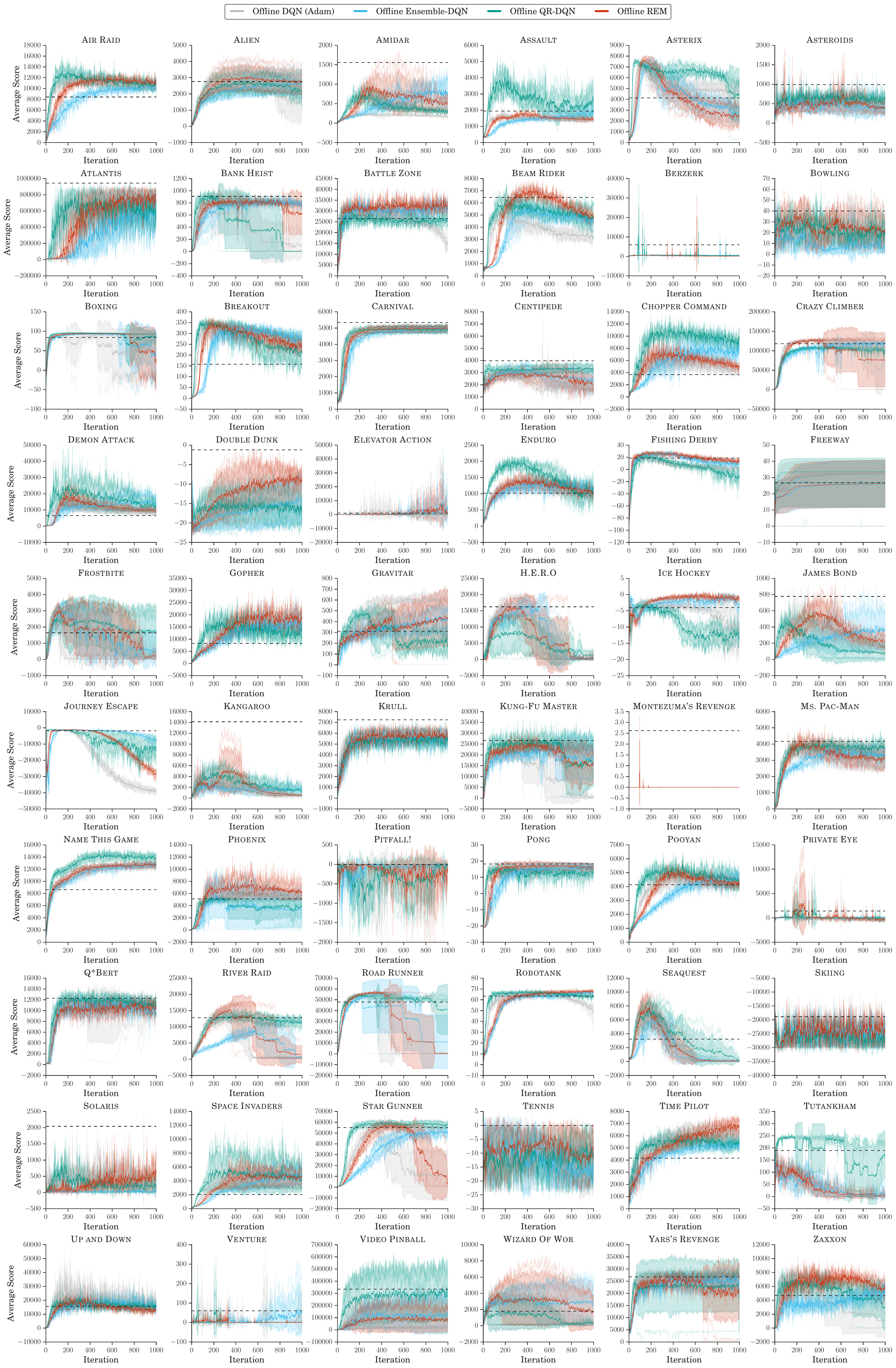}
\vspace{-0.3cm}
\caption{Average evaluation scores across stochastic version of 60 Atari 2600 games of DQN~(Adam), Ensemble-DQN, QR-DQN and REM agents
trained offline using the DQN replay dataset. The horizontal
line for online DQN show the best evaluation performance it obtains during training. All the offline agents except
DQN use the same multi-head architecture with $K=200$ heads. The scores are averaged over 5 runs~(shown as traces)
and smoothed over a sliding window of 5 iterations and error bands show standard deviation.}
\label{fig:batch_agents}
\end{figure*}

\begin{figure*}[t]
\centering
\includegraphics[width=0.90\linewidth]{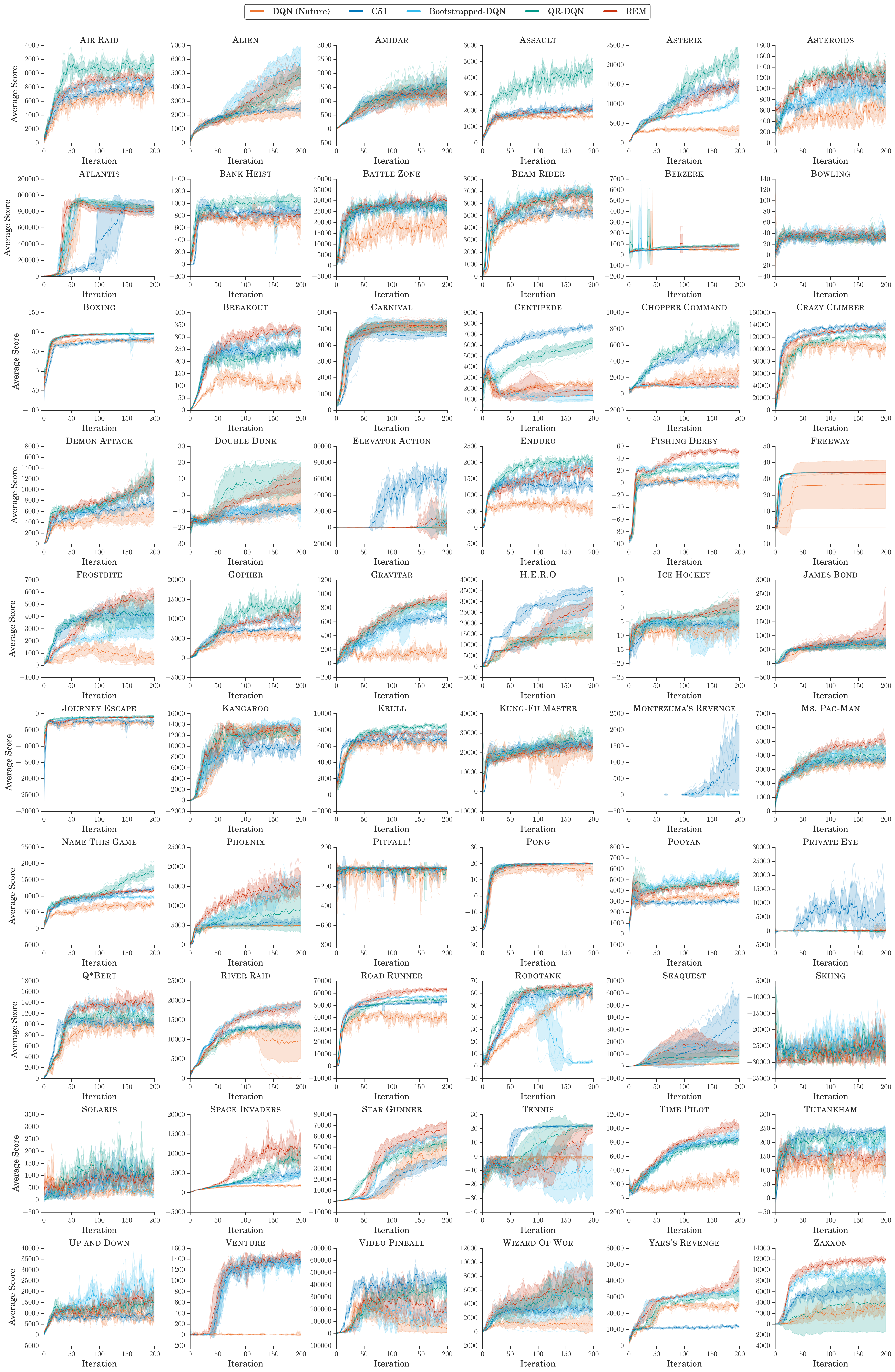}
\caption{\textbf{Online results.} Average evaluation scores across stochastic version of 60 Atari 2600 games of DQN, C51, QR-DQN, Bootstrapped-DQN and REM agents trained online for 200 million game frames~(standard protocol). The scores are averaged over 5 runs~(shown as traces) and smoothed over a sliding window of 5 iterations and error bands show standard deviation.}
\label{fig:online_agents}
\end{figure*}

\begin{figure*}[t]
\centering
\includegraphics[width=0.90\linewidth]{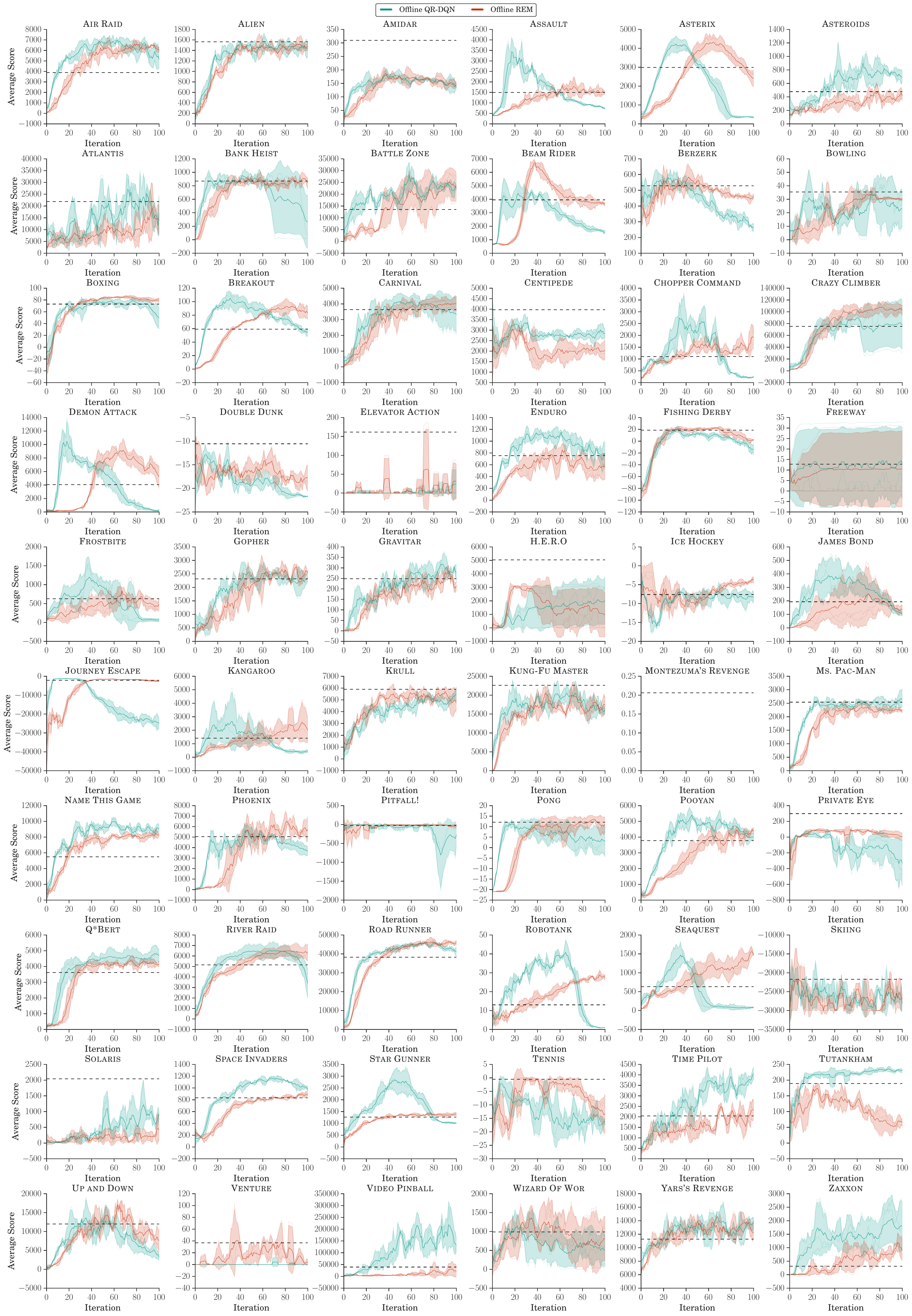}
\vspace{-0.3cm}
\caption{{\bf Effect of Dataset Quality.} Normalized scores~(averaged over 3 runs) of QR-DQN and multi-head REM trained offline on stochastic version of 60 Atari 2600 games for 5X gradient steps using logged data from online DQN trained only for 20M frames~(20 iterations). The horizontal line shows the performance of best policy found during DQN training for 20M frames which is significantly worse than fully-trained DQN.}
\label{fig:random_data_offline_all}
\end{figure*}

\end{document}